%% file: arxiv.tex
\documentclass{article}

\usepackage{wrapfig}
\usepackage[utf8]{inputenc} % allow utf-8 input
\usepackage[T1]{fontenc}    % use 8-bit T1 fonts
\usepackage{hyperref}       % hyperlinks
\usepackage{url}            % simple URL typesetting
\usepackage{booktabs}       % professional-quality tables
\usepackage{amsmath, amssymb, amsthm, amsfonts}       % blackboard math symbols
\usepackage{nicefrac, xfrac}       % compact symbols for 1/2, etc.
\usepackage{microtype}      % microtypography
\usepackage{xcolor}         % colors
\usepackage{bm}
\usepackage{enumitem}
\usepackage{graphicx}
\usepackage{algorithm}
\usepackage{algorithmic}
\usepackage{geometry}
\usepackage{authblk,textcomp}
\usepackage{mathtools}
\usepackage{commath}
\usepackage{physics}
\usepackage[cal=euler]{mathalfa}
\usepackage{libertine}
\usepackage[capitalize,noabbrev]{cleveref}
\usepackage{mathrsfs}
\usepackage{enumitem}
\newlist{condenum}{enumerate}{1} 
\setlist[condenum]{label=(\roman*), ref=(\roman*)}
\crefname{condenumi}{Assumption}{Assumptions}

\theoremstyle{plain}
\newtheorem{theorem}{Theorem}[section]
\theoremstyle{plain}

\theoremstyle{plain}
\newtheorem{result}{Result}[section]

\theoremstyle{definition}
\newtheorem{definition}[theorem]{Definition}
\newtheorem{assumption}[theorem]{Assumption}
\theoremstyle{remark}

\theoremstyle{definition}

\counterwithin{conjecture}{result}

\input{macros}

\hypersetup{pdfauthor={},pdftitle={},%
            colorlinks, linktocpage=true, pdfstartpage=1, pdfstartview=FitV,%
    breaklinks=true, pdfpagemode=UseNone, pageanchor=true, pdfpagemode=UseOutlines,%
    plainpages=false, bookmarksnumbered, bookmarksopen=true, bookmarksopenlevel=1,%
    hypertexnames=true, pdfhighlight=/O,%
    urlcolor=orange, linkcolor=blue, citecolor=blue
        }

\title{Asymptotics of feature learning in two-layer networks\\ after one gradient-step}

\author[1]{Hugo Cui}
\author[2]{Luca Pesce}
\author[2,1]{Yatin Dandi}
\author[2]{Florent Krzakala}
\author[3]{Yue M. Lu}
\author[1]{\\Lenka Zdeborov\'a}
\author[4]{Bruno Loureiro}

\affil[1]{\small Statistical Physics Of Computation lab., 
\'Ecole Polytechnique F\'ed\'erale de Lausanne (EPFL), 1015 Lausanne, Switzerland}
\affil[2]{\small Information Learning \& Physics lab.,
\'Ecole Polytechnique F\'ed\'erale de Lausanne (EPFL), 1015 Lausanne, Switzerland}
\affil[3]{\small Harvard University, 
School of Engineering and Applied Sciences}
\affil[4]{\small D\'epartement d'Informatique, \'Ecole Normale Sup\'erieure (ENS) - PSL \& CNRS, 
F-75230 Paris cedex 05, France}
\makeatletter
\newtheorem*{rep@theorem}{\rep@title}
\newcommand{\newreptheorem}[2]{%
\newenvironment{rep#1}[1]{%
 \def\rep@title{#2 \ref{##1}}%
 \begin{rep@theorem}}%
 {\end{rep@theorem}}}
\makeatother

\theoremstyle{plain}
\numberwithin{theorem}{section}
\newreptheorem{theorem}{Theorem}

\theoremstyle{remark}
\date{\today}

\begin{document}
\date{}
\maketitle

%%%%%%%%%%%%%%%%%%%%%%%%%%%%%%%%%%%%%%%%%%%%%%%%%%%%%%%%%%%%%%%%%%%%%%%%%%%%%%%
\begin{abstract}
In this manuscript, we investigate the problem of how two-layer neural networks learn features from data, and improve over the kernel regime, after being trained with a single gradient descent step. Leveraging the insight from \cite{ba2022high}, we model the trained network by a spiked Random Features (sRF) model. Further building on recent progress on Gaussian universality \cite{dandi2023twolayer}, we provide an exact asymptotic description of the generalization error of the sRF in the high-dimensional limit where the number of samples, the width, and the input dimension grow at a proportional rate. The resulting characterization for sRFs also captures closely the learning curves of the original network model. This enables us to understand how adapting to the data is crucial for the network to efficiently learn non-linear functions in the direction of the gradient -- where at initialization it can only express linear functions in this regime.
\end{abstract}

%%%%%%%%%%%%%%%%%%%%%%%%%%%%%%%%%%%%%%%%%%%%%%%%%%%%%%%%%%%%%%%%%%%%%%%%%%%%%%%
% MAIN
%%%%%%%%%%%%%%%%%%%%%%%%%%%%%%%%%%%%%%%%%%%%%%%%%%%%%%%%%%%%%%%%%%%%%%%%%%%%%%%

\input{sections/intro.tex}
\input{sections/motivation.tex}

\input{sections/main_results}

\input{sections/Derivation}

\input{sections/acknowledgements}

\bibliographystyle{unsrt}
\bibliography{biblio}

%%%%%%%%%%%%%%%%%%%%%%%%%%%%%%%%%%%%%%%%%%%%%%%%%%%%%%%%%%%%%%%%%%%%%%%%%%%%%%%
% Appendix
%%%%%%%%%%%%%%%%%%%%%%%%%%%%%%%%%%%%%%%%%%%%%%%%%%%%%%%%%%%%%%%%%%%%%%%%%%%%%%%
\appendix
\input{sections/appendix/mapping}
\input{sections/appendix/cGET}
\input{sections/appendix/replica.tex}
\input{sections/appendix/non_uniform}

\input{sections/appendix/Bounds}

\input{sections/appendix/Spectrum}
\end{document}

%% file: macros.tex
\newcommand{\prox}{\mathrm{prox}}

\renewcommand{\vec}[1]{\boldsymbol{#1}}

  \makeatletter
  \def\sign{\@ifnextchar*{\@sgnargscaled}{\@ifnextchar[{\sgnargscaleas}{\@ifnextchar{\bgroup}{\@sgnarg}{\sgn} }}}
  \def\@sgnarg#1{\sgn\rbr{#1}}
  \def\@sgnargscaled#1{\sgn\rbr*{#1}}
  \def\@sgnargscaleas[#1]#2{\sgn\rbr[#1]{#2}}
  \makeatother

  \providecommand{\W}{W}


\newcommand{\speedup}[1]{{\color{gray}(\ifdim #1 pt > 0.3pt #1\else $< #1$\fi{}$\times$)}}
\usepackage{amsfonts}
\usepackage{amsmath}
\usepackage{amsthm}
\usepackage{amssymb}
\usepackage{dsfont}

\usepackage{xcolor}
\usepackage{color}
\usepackage{graphicx}

\usepackage{verbatim}
\usepackage{xspace} %

\usepackage{enumerate}
\usepackage{enumitem}

\makeatletter
\newsavebox{\@brx}
\newcommand{\llangle}[1][]{\savebox{\@brx}{\(\m@th{#1\langle}\)}%
  \mathopen{\copy\@brx\mkern2mu\kern-0.9\wd\@brx\usebox{\@brx}}}
\newcommand{\rrangle}[1][]{\savebox{\@brx}{\(\m@th{#1\rangle}\)}%
  \mathclose{\copy\@brx\mkern2mu\kern-0.9\wd\@brx\usebox{\@brx}}}
\makeatother

\DeclareMathOperator{\Ea}{\mathbb{E}}

  \usepackage{bm}

\providecommand{\mycomment}[3]{\todo[caption={},size=footnotesize,color=#1!20]{\textbf{#2: }#3}}%
\providecommand{\inlinecomment}[3]{%
  {\color{#1}#2: #3}}%
\newcommand\commenter[2]%
{%
  \expandafter\newcommand\csname i#1\endcsname[1]{\inlinecomment{#2}{#1}{##1}}
  \expandafter\newcommand\csname #1\endcsname[1]{\mycomment{#2}{#1}{##1}}
}

  \definecolor{mydarkblue}{rgb}{0,0.08,0.45}
  \usepackage{hyperref}
  \hypersetup{ %
    pdftitle={},
    pdfauthor={},
    pdfsubject={},
    pdfkeywords={},
    pdfborder=0 0 0,
    pdfpagemode=UseNone,
    colorlinks=true,
    linkcolor=mydarkblue,
    citecolor=mydarkblue,
    filecolor=mydarkblue,
    urlcolor=mydarkblue,
    pdfview=FitH}

\usepackage{url}

%% file: sections/intro.tex
\section{Introduction}
\label{sec:main:introduction}
A common deep learning intuition behind the unreasonable effectiveness of neural networks is their capacity to effectively adapt to the training data which makes them superior to kernel methods. While kernel methods and their finite width approximations are known to be data-hungry (see e.g. \cite{adlam2023kernel}), neural networks have proven themselves to be flexible and efficient in practice. Their adaptivity and capacity to learn features from data are behind the success in efficiently solving problems from image classification to text generation. A large part of our current theoretical understanding of neural networks stems from the investigation of their lazy regime where features are {\it not} learned during training. This includes a set of works that fall under the umbrella of Gaussian processes \cite{neal1996bayesian, lee2018deep}, the Neural Tangent Kernel (NTK) \cite{jacot2018neural} and the Lazy regime \cite{chizat_2019_lazy}. A crucial question in the theoretical machine learning community is thus to characterize the advantages of two-layer neural networks beyond these convex optimization approaches. \\

Different theoretical works have offered sharp separation results between kernel and feature learning regimes (see e.g. \cite{Ghorbani2019,refinetti2021classifying,damian2022neural,abbe2023sgd, shi2022theoretical}). In particular, \cite{ba2022high,damian2022neural}, and later \cite{dandi2023twolayer}, discussed the advantage of neural networks when training with only {\it one single step} of large-batch gradient descent with a large learning rate. Specifically, \cite{ba2022high} highlighted that the weight matrix after the first training step can be decomposed in a bulk plus a rank-one spike, effectively mapping the learned features to a \emph{spiked Random Features model} (sRF), defined in eq.~\ref{eq:def:sRF}). This observation has fueled many further studies on the effect of spiked structure, see e.g. \cite{damian2022neural,dandi2023twolayer, ba2023learning, mousavi2023gradient, moniri2023theory}. \looseness=-1\\

In this paper, we follow this line of work and provide an exact high-dimensional description of the test error achieved by a sRF, which we use to model a two-layer network after a {\it single, large, gradient step}. Our work provides a sharp asymptotic treatment of a setting where feature learning is modeled in a non-perturbative, high-dimensional regime, with a model able to express non-linear functions beyond polynomials. This analysis quantitatively illustrates the benefits of feature learning over the lazy regime.\looseness=-1\\

Our \textbf{main contributions} in this paper are the following: 

\begin{itemize}
    \item \textbf{Exact asymptotics for sRF --} We provide a sharp asymptotic characterization of the test error, alongside a set of summary statistics, for a sRF. We discuss analytically, and provide numerical support, why sRFs constitute good approximations for two-layer neural networks with first layer weights trained with a large learning rate gradient step. The derivation leverages the (non-rigorous) replica method from statistical physics \cite{parisi1979toward,parisi1983order}, and provides a set of scalar self-consistent equations for the generalization error.
    \item \textbf{Conditional Gaussian Equivalence --} Building upon \cite{dandi2023twolayer}, we show (and provide strong numerical evidence to support) that the learning properties of the sRF model are asymptotically equivalent to a simple \emph{conditional} Gaussian model in the high-dimensional proportional regime. The conditional Gaussian distribution is characterized by the projections of the input data on the spike in the weight matrix. This mapping constitutes the extension of related theoretical results that unveiled a similar Gaussian equivalence property for the training and generalization error for non-spiked vanilla RFs \cite{goldt_gaussian_2021,goldt2020modeling,gerace_generalisation_2020,hu2022universality,Dhifallah2020,mei_generalization_2022, cui2023bayes,schroder2023deterministic,bosch2023precise,dandi2023universality}. \looseness=-1
  \item \textbf{Feature learning --} We provide an extensive discussion on how feature learning leads to a drastic improvement in the generalization performance over random features in a data-limited regime, demonstrating a clear and quantitative separation with respect to kernel method and random feature models. In particular, we derive both upper and lower bounds on the generalization error and discuss under which conditions they are tight.
\end{itemize}

The code used in the present work is available \href{https://github.com/SPOC-group/OneStepGD_asymptotics}{in this repository.}\looseness=-1

%% file: sections/motivation.tex
\section{Setting, Motivation and Related Work}
\label{sec:motivation}
\paragraph{Setting ---} We study fully-connected two-layer networks
\begin{align}
\label{motiv:student}
f_{W,\vec a}(\vec x)=\frac{1}{\sqrt{p}}\sum\limits_{i=1}^{p} a_{i} \sigma(\vec{w}_{i}^{\top}\vec x)\,,
\end{align}
and their capacity to learn a single-index target function of isotropic Gaussian covariates:
\begin{align}
\label{eq:def:target}
    f_{\star}(\vec x) = \sigma_{\star}(\sfrac{\vec{\theta}^{\top}\vec{x}}{\sqrt{d}}), \quad \vec{x}\sim\mathcal{N}(0,I_{d})
\end{align}
from finite batch ${\mathcal{D}=\{(\vec{x}^{\mu},y^{\mu})_{\mu=1}^{n}\}}$ of $n$ independently drawn training samples. While we do not consider bias terms for conciseness, we discuss in Appendix \ref{sec:app:replica} how such terms can be treated. We further assume the activation $\sigma$ to be odd. We consider a layer-wise training procedure where the first layer weights $W\in\mathbb{R}^{p\times d}$ are trained for a single gradient step:

\begin{align}
\label{eq:def:gradstep}
    &\vec{w}^{(1)}_{i} = \vec{w}^{(0)}_{i} - \eta \vec{g}_{i}\\
    &\vec{g}_{i} = \frac{1}{\sqrt{p}n_0}\sum\limits_{\mu=1}^{n_{0}}(y^{\mu}-f_{W^{(0)},\vec{a}^{(0)}}(\vec{x}^{\mu}))a_{i}^{(0)}\sigma'(\vec{w}^{(0) \top}_{i}\vec{x}^{\mu})\vec{x}^{\mu}\notag
\end{align}
on a subset $\mathcal{D}_{0}\subset\mathcal{D}$ of size $n_{0}$, where $(W^{(0)}, \vec{a}^{(0)})$ denote the initial weights and $\eta\!>\!0$ the learning rate. For simplicity, we assume $\vec{a}^{(0)}\!=\! \sfrac{\vec{1}_{p}}{\sqrt{p}}$ (uniform initialization) and $\vec{w}_{i}^{(0)}$ with unit norm $\lVert \vec{w}^{(0)}_{i}\lVert =\! 1$ and weak correlation ${\vec{w}^{(0)}_{i}}\cdot\vec{w}^{(0)}_{j} =O_{d}(\sfrac{\mathrm{polylog}(d)}{\sqrt{d}})$ for $i\neq j$ (e.g. uniformly drawn from the unit sphere $\mathbb{S}^{d-1}$). Finally, we assume that the second layer initialization is not informed, in the sense that its overlap with the target weights $\vec{\theta}$ is asymptotically small, $\sfrac{\vec{a}^{(0)}\cdot \vec{\theta}}{\sqrt{p}}=O_d(\sfrac{1}{\sqrt{d}})$. Given the updated weights $W^{(1)}$, we train the read-out layer on the remaining data $\mathcal{D}_{1} \!=\! \mathcal{D}\setminus\mathcal{D}_{0}$: 
\begin{align}
\label{eq:ERM_a}
    \hat{\vec a}_{\lambda} \!=\! \underset{\vec a\in\mathbb{R}^{p}}{\mathrm{argmin}}\frac{1}{2}\sum\limits_{\mu=1}^{n_{1}}\left(y^{\mu}-f_{W^{(1)},\vec a}(\vec x^{\mu})\right)^2 \!+\! \frac{\lambda}{2}\lVert \vec a\lVert^2.
\end{align}
with $\lambda \in\mathbb{R}_{+}$ being a regularization parameter. Note that the layer-wise training procedure considered here is commonly studied in the theoretical machine learning literature \cite{ba2022high, damian2022neural,abbe2023sgd, berthier2023learning, dandi2023twolayer, moniri2023theory} due to its mathematical tractability.\\

Our main goal in the following is to describe the generalization error:
\begin{align}
\label{eq:mse}
    \epsilon_g=\mathbb{E}_{\mathcal{D},\vec x}\left(f_\star(\vec x)-f_{W^{(1)}, \hat{\vec{a}}_{\lambda}}(\vec x)\right)^2.
\end{align}
in the high-dimensional proportional limit where $n_0,n_1,d,p,\eta\!\to\!\infty$ at fixed ratios $\alpha_{0} \!=\! \sfrac{n_{0}}{d}$, ${\alpha\!=\!\sfrac{n_{1}}{d}},\beta \!=\! \sfrac{p}{d}, \Tilde{\eta}\!=\!\sfrac{\eta}{d}$. \looseness=-1

\paragraph{Motivation ---} Driven by the lazy-training regime of learning of large-width networks \cite{chizat_2019_lazy}, a large body of literature has been dedicated to the particular case where the first layer weights are fixed at initialization $W^{(0)}$ ($\eta=0$), also known as the Random Features (RF) model \cite{rahimi2007random}. In particular, \cite{Ghorbani2019, Ghorbani2020, Mei2023} have shown that with $n_{1}=\Theta_{d}(d)$ samples $f_{W^{(0)},\vec\hat{\vec{a}}_{\lambda}}$ can only approximate, at best, a linear function of $\vec{x}^{\mu}$, with the non-linear part playing a role akin to additive label noise. This is a strong limitation: RF network requires {\it polynomials} number of data and neurons to fit a simple polynomial \cite{Mei2023,xiao2022precise}. It is one of our motivations here to discuss how, with a single gradient step, these limitations are lifted.\looseness=-1\\

Behind the scenes in this effective linearity of random features is a Gaussian Equivalent Principle (GEP) \cite{goldt_gaussian_2021,mei_generalization_2022,hu2022universality,montanari2022universality, dandi2023universality}, which states that in this regime the random feature map $\varphi = \sigma(W^{(0)}\vec{x})$ is statistically equivalent to a rescaled stochastic linear map $\varphi^{g}(\vec{x})\asymp \mu_{0}\vec{1}+\mu_{1}W^{(0)}\vec{x}+\mu_{\star}\vec{z}$, with $\vec{z}\sim\mathcal{N}(0,I_{p})$. This surprising universality result allows to go beyond lower bounds for the generalization performance, making the problem amenable to a tight high-dimensional characterization of all relevant statistics in these models  \cite{mei_generalization_2022, gerace_generalisation_2020, Dhifallah2020}. \\

\begin{wrapfigure}{r}{0.5\textwidth}
    \centering
    \includegraphics[scale=0.66]{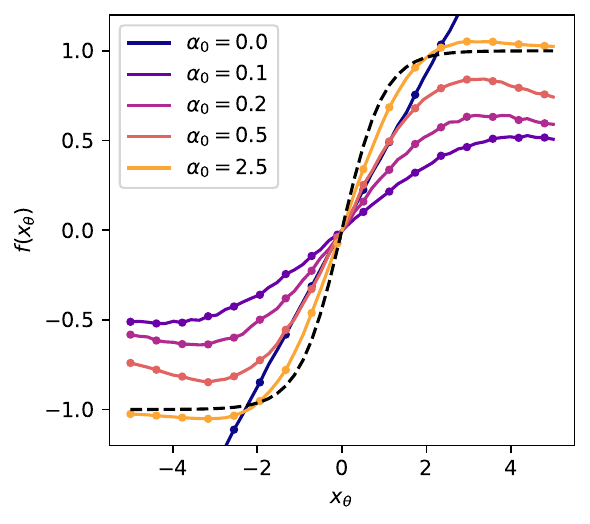}
    \caption{Numerical estimation of the function $f(x_\theta)=\mathbb{E}_x[f_{W^{(1), \hat{\vec{a}}_\lambda}}|\sfrac{\vec \theta^\top \vec x}{\sqrt{d}}=x_\theta]$ implemented by the trained network~\eqref{motiv:student} in the direction spanned by the weights $\vec \theta$ of the target function. The activations are $\sigma=\sigma_\star=\tanh$, and simulations were run in dimensions $d=p=2000$, for a learning rate $\eta=2.5 p$, and a readout regularization $\lambda=0.01$. The readout was trained with $n_1=2d$ samples. Different colors corresponds to different sample complexities $\alpha_0\equiv\sfrac{n_0}{d}$ used to implement the gradient step on the first layer weights, with $\alpha_0=0$ corresponding to not implementing the step.}
    \label{fig:NL}
\end{wrapfigure}

Fig.~\ref{fig:NL} illustrates this fundamental limitation of RF models contrasted with the function $f(x_\theta)\!=\!\mathbb{E}_{\vec x}[f_{W^{(1)},\vec \hat a)\lambda}(\vec x)|\sfrac{\vec \theta^\top \vec x}{\sqrt{d}}=x_\theta]$, along the direction of the target  $\vec \theta$, implemented by the network $f_{W^{(1)}, \vec a}$ trained with a single gradient step (eq.~\eqref{eq:def:gradstep}). Varying the amounts of data $n_0\!=\!\alpha_0 d$ used in the first gradient step, the function $f(x_\theta)$ moves from a linear approximation of $\sigma_{\star}$ in the RF limit ($\alpha_0 \!=\! 0$), to an accurate non-linear one ($\alpha_0 \!=\! 2.5$).\\

Learning a non-linear approximation of $\sigma_{\star}$ in the high-dimensional proportional regime therefore requires learning features. \cite{ba2022high} have proven that with $\eta=\Theta_{d}(1)$, the GEP holds even after a few gradient steps, corroborating a fact empirically observed by \cite{loureiro_learning_2021}. Indeed, they have shown that $\eta = \Theta_{d}(d)$ is \emph{sufficient} to go beyond a linear approximation of $f_{\star}$ in this regime.  \cite{moniri2023theory} considered intermediate scalings of step-size $\eta= \Theta_d(d^s)$ for $\sfrac{1}{2} \!<\! s  \!<\! 1$, which allows the network to fit target functions along $\vec{\theta}$ having finite degree, providing a precise characterization of the train and test errors. In this intermediate regime, the feature matrix can be approximated through a finite number of spikes corresponding to increasing degree of functions along $\vec \theta$. Instead, we consider the full scaling of $\eta = \Theta_{d}(d)$, where such a finite-spike approximation is insufficient and the network can fit arbitrary functions along $\vec \theta$, and provide an approximate characterization through the lens of sRF models. \cite{dandi2023twolayer} proved that even if the target depends on multiple directions (multi-index model), only a (non-linear) function of a {\it single direction} $\vec{\theta}$ can be learned with a single gradient step and $\eta = \Theta_{d}(d)$. This observation justifies the focus on single-index functions \eqref{eq:def:target} on the regime of interest.

\paragraph{Further related works --} \textbf{Random features} were first introduced as a computationally efficient approximation to kernel methods \cite{rahimi2007random}. Recently, they have enjoyed renewed interest also as models of two-layer neural networks in the lazy regime. Tight asymptotics for the random features model have been derived by \cite{goldt2020modeling, goldt_gaussian_2021, gerace_generalisation_2020, mei_generalization_2022, hu2022universality, Dhifallah2020} in the two-layer case, and were extended to deep networks in \cite{schroder2023deterministic, schroder2024asymptotics, bosch2023precise} in the deep case. Importantly, with the exception of \cite{gerace_generalisation_2020} who considered rotationally invariant weights and \cite{zavatone2023learning} for the case of deep \textit{linear} random features, all these works assumed unstructured weights. In sharp contrast, gradient-trained neural networks have fundamentally structured weights. In the present manuscript, we consider such a case, by analyzing sRF models, for which the weights are given by a bulk random matrix plus a rank-one spike.\looseness=-1\\

% \paragraph{Feature learning regime --} 
\noindent\textbf{Feature learning regime --} Perturbative feature learning corrections to the large-width lazy regime have been extensively studied in the literature \cite{pmlr-v107-yaida20a, Hanin2020Finite, Dyer2020Asymptotics, Seroussi2023, NEURIPS2021_b24d2101}. Our work radically contrasts with this line, since we account for feature learning in the first step, \emph{non-perturbatively} (note the gradient in \eqref{eq:def:gradstep} has a norm comparable with the initial weights). Beyond the lazy regime, a major recent development has been the understanding that the training dynamics of two-layer neural networks with small learning rates can be mapped to a Wasserstein gradient flow, known as the \textit{mean-field regime} \cite{mei2018mean,chizat2018global,rotskoff2018trainability,sirignano2020mean,bordelon2023dynamics}. Over the past few years, this flow was investigated under different classes of generative data models, such as staircase functions \cite{abbe2021staircase, abbe2022merged, abbe2023sgd}, single-index \cite{berthier2023learning, arnaboldi2023escaping} and multi-index models \cite{arnaboldi.stephan.ea_2023_high}, symmetric targets \cite{hajjar2023symmetries} and Gaussian mixture models \cite{pmlr-v139-refinetti21b, ben2022high}.
\looseness=-1

%% file: sections/main_results.tex
\section{Main Technical Results}
\label{sec:setting}
Our main technical results are a tight asymptotic characterization of the test error achieved by sRFs (defined in \ref{eq:def:sRF}), which we argue provide good models for two-layer networks trained with a single large gradient step followed by a ridge regression on the readout weights. We provide compelling numerical support that this theoretical characterization captures very closely the learning curves of two-layer networks trained following the protocol \eqref{eq:def:gradstep}, and that sRF thus provide a valuable analytical playground to understand the learning and behaviour of the latter. These results are enabled through the expression of the parameters of an equivalent sRF model in terms of the training parameters of the original network in subsection \ref{subsec:sRF_mapping}, which can in turn be mapped to an equivalent Gaussian model in subsection \ref{subsec:cGET}. Subsection \ref{subsec:test_error} finally states the tight asymptotic characterization of the test error.\looseness=-1

\subsection{Asymptotics of the first layer weights after one (large) gradient step}
\label{subsec:sRF_mapping}
The first step is to derive an explicit asymptotic expression for the hidden-layer weights $W^{(1)}$ after one (large) gradient step. In the following, we show that the learning problem introduced in Section \ref{sec:motivation} can be modelled by a \emph{spiked Random Features} model, which we first define.

\begin{definition}[sRF model]
    We define a \emph{spiked Random Features} (sRF) model with bulk variance $c$, spike strength $r$ as the two-layer neural network 
    \begin{align}
\label{eq:def:sRF}
g_{F, \vec{a}}(x)= \frac{1}{\sqrt{p}}\vec{a}^{\top}\sigma\left(Fx\right)
\end{align}
    with trainable readout $\vec a$ and frozen random first layer weights:
   \begin{align}
    \label{eq:W}
   F=\W+r\frac{\vec u \vec v^\top }{\sqrt{d}}.
\end{align}
where $\W$ is a random matrix with rows independently sampled from $\mathbb{S}^{d-1}(\sqrt{c})$, and $\vec{u},\vec{v}\in\mathbb{S}^{d-1}(\sqrt{d})$. We further say that a sRF has alignment $\gamma$ with $\vec \theta$ when $\vec v$ is uniformly sampled uniformly sampled among vectors with norm $\sqrt{d}$ satisfying $\sfrac{\vec v^\top \vec\theta}{d}=\gamma$.

\end{definition}

As discussed above, after a large gradient step, the first layer weights $W^{(1)}$ take the form \cite{ba2022high, dandi2023twolayer}:
\begin{align}
\label{eq: decompo}W^{(1)}=W^{(0)}+\Delta+r\frac{\check{\vec{u}}\check{\vec{v}}^\top }{\sqrt{d}},
\end{align}
where the precise expressions of $\Delta, \check{\vec{u}},\check{\vec{v}}$ are detailed in Appendix \ref{app:GD}. It has thus
developed spikes $\check{\vec{u}}$ and $\check{\vec{v}}$ that correlated to the target weights $\vec{\theta}$, while the bulk, originally $W^{(0)}$, also gets modified by an additional term $\Delta$ and displays as a consequence a rescaled variance. These changes make it reasonable to model the two-layer neural network with a sRF with matching parameters $c,r,\gamma$. The expression of these parameters depend on the specifications $\tilde{\eta},\alpha_0,\beta, \sigma,\sigma_\star$ of the network and training protocol. Expliciting these expressions is the object of the following result. 
\begin{result}
[Equivalent sRF model]
\label{res:GD_mapping}
Consider two-layer networks with first-layer weights trained with a single gradient step of learning rate $\eta$ from initial conditions $\vec{a}^{(0)}=\sfrac{\vec{1}_{p}}{\sqrt{p}}$ and $\vec{w}_{i}^{(0)}$ with unit norm $\lVert \vec{w}^{(0)}_{i}\lVert = 1$ and weak correlation ${\vec{w}^{(0)}_{i}}\cdot\vec{w}^{(0)}_{j} =O_{d}(\sfrac{1}{\sqrt{d}})$ for $i\neq j$ (eq.~\eqref{eq:def:gradstep}). In the asymptotic limit $n_0,n,d,p\to \infty$, with $ \alpha_0=\sfrac{n_0}{d},\alpha=\sfrac{n}{d},\beta=\sfrac{p}{d},\tilde{\eta}=\sfrac{\eta}{d}=\Theta_d(1)$, the quantities $c=\sfrac{1}{pd}\lVert W_0+\Delta\lVert^2,~r=\sfrac{\lVert \check{\vec{u}}\lVert \lVert\check{\vec{v}}\lVert}{\sqrt{pd}}, \gamma=\sfrac{\vec{\theta}^\top \check{\vec{v}}}{\lVert\check{\vec{v}}\lVert\lVert\vec\theta\lVert}$ \eqref{eq: decompo} concentrated to the values
\begin{align}
    \label{eq:GD_sRF}
    &c= 1+\frac{\Tilde{\eta}^2\check{h}_1^2 h_2^\star}{\alpha_0 \beta^2}\\   \label{eq:def:r}
    &r=\frac{\Tilde{\eta}h_1}{\beta}\left(\frac{h_2^\star}{\alpha_0}+h_1^{\star 2}\right)^{\sfrac{1}{2}}\\
    &\gamma = {\frac{h_1^\star}{\left(\frac{h_2^\star}{\alpha_0}+h_1^{\star 2}\right)^{\sfrac{1}{2}}}}
\end{align}
where:
\begin{align}
    &h_1=\mathbb{E}_z[z\sigma(z)], &&h^\star_1=\mathbb{E}_z[z\sigma_\star(z)],\notag\\
    &\check{h}_1^2=\mathbb{E}_z[(\sigma^\prime(z)-h_1)^2], &&
    h^\star_2=\mathbb{E}_z[\sigma_\star(z)^2],
\end{align}
with $z\sim\mathcal{N}(0,1)$. 
These parameters $c,r,\gamma$ specify an \textit{equivalent sRF}, which we take as a model of the network after one gradient step.
\end{result}
The derivation of Result \eqref{res:GD_mapping} is detailed in Appendix \ref{app:GD}. One may wonder what are the sources of approximation when modeling two-layer networks after a gradient step \eqref{eq:def:gradstep} by a sRF \ref{eq:def:sRF}. In considering the equivalent sRF, one is effectively modeling the bulk term $\Delta$ in \eqref{eq: decompo} by a random matrix with independent rows sampled from the sphere, with matching norm. This approximation is reasonable in view of the results of \cite{dandi2023twolayer} (Lemma $12$), who rigorously establish the asymptotic near-orthogonality of the rows of $\Delta$, and that they have vanishing overlap with $\vec{\theta}$. We are, on the other hand, ignoring possible non-trivial structures in the row covariance of $\Delta$. However, as we shall discuss later, for all probed usual activation functions and setups, the sRF provides a quantitatively close approximation. Let us further stress that beyond being approximations of two-layer networks after a gradient step, sRF models constitute an important and natural generalization of the celebrated and extensively probed RF models \cite{rahimi2007random, gerace_generalisation_2020, mei_generalization_2022, hu2022universality}, which crucially allow to learn \textit{non-linear} functions --  where RFs can only express linear functions. They hence are also models of standalone interest.
Finally, note that above we have assumed a uniform initialization for the readout layer. This can be relaxed in the equivalence above, for instance for $\vec{a}^{(0)}$ taking a finite number of values, leading to a finite-rank term instead of a single spike. The emergence or usefulness of low-rank structures in network weights has been observed and discussed in numerous empirical and theoretical works, e.g. \cite{hu2021lora, allen2022feature, wang2022spectral}.

\subsection{Conditional Gaussian equivalence}
\label{subsec:cGET}
The sharp characterization of the test performance, which we state in Result~\ref{res:asymptotics} in the following subsection, is enabled by further mapping the sRF model to an exactly solvable (conditional) Gaussian model. We adapt the rigorous result in \cite{dandi2023twolayer} (Theorem~{\bf 4}, second point) by constructing explicitly the equivalent stochastic feature map, that we believe is of independent interest.
\begin{result}[Conditional Gaussian Equivalence]
\label{conj:main:gaussian_equivalence}
Consider the sRF model with weights $F=\W + r\sfrac{\vec u \vec v^\top}{\sqrt{d}}$, with $\vec{u}=\vec{1}_p$ and parameters $c,r,\gamma$, and the corresponding feature map given by
\begin{align}
\label{eq:phi}
    \vec \varphi (\vec x)= \sigma \left( F \vec x \right)
\end{align}

Define the equivalent stochastic feature map  
\begin{align}
\label{eq:phig}
    \vec \varphi^g(\vec x)\overset{d}{=}\mu_0(\kappa) \vec 1_p+\mu_1(\kappa)\W\vec x+\mu_2(\kappa)\mathcal{N}(0, \mathbb{I}_p),
\end{align}
where $\kappa \equiv \sfrac{\vec v^\top \vec x}{\sqrt{d}}$. We introduced the coefficients $\mu_0(\kappa),\mu_1(\kappa),\mu_2(\kappa)$ defined as
\begin{align}
\label{eq:main:mus}
&\mu_0(\kappa) =  \Ea_{z}{\sigma(z+ r \kappa)} , \notag\\&
\mu_1(\kappa)=  \frac{1}{c}\Ea_{z}{z\sigma(z+r \kappa)},    \notag\\
&\mu_2(\kappa) = \sqrt{\Ea_{z}{\sigma^2(z+ r \kappa)-c(\mu_1(\kappa))^2-(\mu_0(\kappa))^2}},
\end{align}
with expectations bearing over $z\sim\mathcal{N}(0,c)$. 
%The test error achieved by performing ridge regression on the sRF feature $\vec \varphi$ \eqref{eq:phi} by the equivalent feature map $\vec \varphi^g$ \eqref{eq:phig} in \eqref{eq:mse} leads to asymptotically identical test errors $\epsilon_g$.
The test error $\epsilon_g$ achieved by ridge regression
\begin{align}
\label{eq:ERM_a_cGET}
    \hat{\vec a}_{\lambda} = \underset{\vec a\in\mathbb{R}^{p}}{\mathrm{argmin}}\frac{1}{2}\sum\limits_{\mu=1}^{n_{1}}\left(y^{\mu}-\frac{1}{\sqrt{p}}\vec a^\top \phi(\vec x) \right)^2 + \frac{\lambda}{2}\lVert \vec a\lVert^2.
\end{align}
is asymptotically identical for $\phi=\varphi$ and $\phi=\varphi^g$.
\end{result}

%We refer to next section for the technical details on the mapping.
Result \ref{conj:main:gaussian_equivalence} extends the similar linearizations provided e.g. in \cite{goldt2020modeling,hu2022universality,cui2023bayes} for unstructured RFs to sRFs \eqref{eq:W}. Informally, the quantity $\kappa$ in the stochastic feature map~\eqref{eq:phig} represents the projection of the input on the spike defining the sRF $\kappa = \sfrac{\vec x^\top \vec v}{\sqrt{d}}$. The equivalent network $\sfrac{1}{\sqrt{d}}\vec a^\top \vec \varphi^g(x)$ obtained by replacing $\vec \varphi$ by the equivalent $\vec \varphi^g$ is a linear combination of terms such as $\mu_0(\kappa)$, $\mu_1(\kappa)\kappa$, $\mu_1(\kappa)\vec a^\top W (\Pi^\perp \vec x)$, plus noise. On an intuitive level, this makes it apparent that sRFs can thus express \textit{non-linear} functions of the component $\kappa$ along the spike $\vec v$, but only linear functions of the component $\Pi^\perp \vec x$ orthogonal thereto. This feature learning correction to the linear regime is visually exemplified in Fig.~\ref{fig:NL}. In the next subsection, we make this discussion more quantitative by providing a tight asymptotic characterization of the test error achieved by two-layer networks trained with a single large gradient step followed by a ridge regression on the readout weights.\looseness=-1

\subsection{Tight asymptotic characterization of the test error}

\label{subsec:test_error}

Finally, we leverage on the sequential mappings of Result~\ref{res:GD_mapping} and Result \ref{conj:main:gaussian_equivalence} to offer sharp asymptotic guarantees on the test error achieved after training the readout weights using eq.~\eqref{eq:ERM_a}.\looseness=-1 

\begin{assumption}
\label{ass:assumption}
    Denote by $\{\boldsymbol{e}_i\}_{i=1}^p$ ($\{\boldsymbol{f}_i\}_{i=1}^d$) the left (resp. right) singular vectors of $\W$. We further note $\{\lambda^\ell_i\}_{i=1}^p$ the squared singular values of $\W$. The squared singular values $\{\lambda^\ell_i\}_{i=1}^p$ and the projection of the teacher vector $\vec \theta$ and the spike $v$ on the eigenvectors $\{\boldsymbol{f}_i^\top \vec v\}_{i,\ell} $,$\{\boldsymbol{f}_i^\top \Pi^\perp \vec \theta\}_{i,\ell} $ are assumed to admit a well-defined joint distribution $\nu$ as $d\to \infty$.\looseness=-1
    \begin{align*}
        &\frac{1}{p}\sum\limits_{i=1}^{\min(p,d)}\delta\left(\lambda_i-\varrho\right)
\delta\left(\boldsymbol{f}_i^\top v-\tau\right)\delta\left(\boldsymbol{f}_i^\top \Pi^\perp \vec \theta-\pi\right
        )\xrightarrow[]{d\to\infty} \nu(\varrho,\tau,\pi).
    \end{align*}
\end{assumption}
%The form of the limiting density $\nu$ is the object of further discussion in Appendix \ref{sec:app:replica}. One is now in a position to state under Assumption \ref{ass:assumption} our main technical result, namely a tight characterization of the test error achieved by two-layer networks trained with a single large gradient step, when the readout weights are subsequently trained by ridge regression.

\begin{result}[Test error asymptotics]
\label{res:asymptotics}
Consider the ERM problem associated with the training of the readout weights $\vec a$ of the equivalent sRF model defined by Result \ref{res:GD_mapping} (namely replacing the network $f_{W^{(1)}\vec a}$ by the equivalent sRF $g_{F, \vec a}$ in \eqref{eq:ERM_a}), and assume~\ref{ass:assumption} to hold. Define $\Pi^\perp\equiv  \mathbb{I}_d-\sfrac{vv^\top}{d}$ the projection to the subspace orthogonal to the spike $\vec v$. In the asymptotic limit $d,p,n\to\infty$ with $\alpha=\sfrac{n}{d},\beta=\sfrac{p}{d}=\Theta_d(1)$, the summary statistics
\begin{align}
    \label{eq:summary_statistics}
    &q_1=\frac{\hat{\vec a}^\top \W\Pi^\perp \W^\top \hat{\vec a}}{p}, && q_2=\frac{\hat{\vec a}^\top \hat{\vec a}}{p}, \notag\\
    ¨&m=\frac{\vec 1_p^\top \hat{\vec a}}{\sqrt{p}}, &&\zeta=\frac{\hat{\vec a}^\top \W \vec v}{\sqrt{dp}}, \notag\\
    &\psi=\frac{\hat{\vec a}^\top \W\Pi^\perp \vec{\theta}}{\sqrt{dp}}, && \rho^2=\frac{\vec \theta^\top \Pi^\perp \vec \theta}{d}
\end{align}
concentrate in probability to the solutions of the system of equations
\begin{align}
\label{eq:main:SP}
    &\begin{cases}
        q_1=\int d\nu(\varrho,\tau,\pi) \varrho\frac{\left(
\hat{q}_1 \varrho +\hat{q}_2
+\hat{\zeta}^2 \varrho\tau^2 +\hat{\psi}^2 \varrho\pi^2
\right)}{
\left(\lambda+\hat{V}_1\varrho +\hat{V}_2\right)^2
}-\beta\hat{\zeta}^2\frac{I(\hat{V}_1, \hat{V}_2)^2}{\left(1-\beta\hat{V}_1I(\hat{V}_1,\hat{V}_2)\right)^2}-\hat{\zeta}^2\frac{\int d\nu(\varrho,\tau,\pi) \frac{\tau^2\varrho^2}{(\lambda+\hat{V}_1\varrho +\hat{V}_2)^2}\left[
    \left( 1-\beta\hat{V}_1I(\hat{V}_1,\hat{V}_2)\right)^2-1
    \right]}{\left(1-\beta\hat{V}_1I(\hat{V}_1,\hat{V}_2)\right)^2}
\\
    q_2=\int d\nu(\varrho,\tau,\pi) \frac{\left(
\hat{q}_1 \varrho +\hat{q}_2
+\hat{\zeta}^2 \varrho\tau^2 +\hat{\psi}^2 \varrho\pi^2
\right)}{
\left(\lambda+\hat{V}_1\varrho +\hat{V}_2\right)^2}-\hat{\zeta}^2\int d\nu(\varrho,\tau,\pi) \frac{\tau^2\varrho}{(\lambda+\hat{V}_1\varrho +\hat{V}_2)^2}\left[
1-\frac{1}{\left(1-\beta\hat{V}_1I(\hat{V}_1,\hat{V}_2)\right)^2}
\right]\\
V_1=\int d\nu(\varrho,\tau,\pi) \varrho\frac{1}{
\lambda+\hat{V}_1\varrho +\hat{V}_2}\\
V_2=\int d\nu(\varrho,\tau,\pi) \frac{1}{
\lambda+\hat{V}_1\varrho +\hat{V}_2}\\
m=\frac{1}{\mathbb{E}_\kappa\left[\frac{\mu_0(\kappa)^2}{1+V(\kappa)}\right]}
\mathbb{E}_{\kappa,y}\left[\frac{\mu_0(\kappa)(\sigma_\star(\kappa, y)-\mu_1(\kappa)\kappa\zeta)}{1+V(\kappa)}\right]\\
\zeta=\hat{\zeta}\sqrt{\beta} \int d\nu(\varrho,\tau,\pi) \varrho\tau^2\frac{1}{
\lambda+\hat{V}_1\varrho +\hat{V}_2} \\
\qquad+\beta^{\sfrac{3}{2}}\hat{\zeta}\hat{V}_1\frac{I(\hat{V}_1,\hat{V}_2)^2}{1-\beta\hat{V}_1I(\hat{V}_1,\hat{V}_2)}\\
\psi=\hat{\psi}\sqrt{\beta} \int d\nu(\varrho,\tau,\pi) \varrho\pi^2\frac{1}{
\lambda+\hat{V}_1\varrho +\hat{V}_2}
    \end{cases}
    \\
&\begin{cases}
\hat{V}_1=\frac{\alpha}{\beta}\mathbb{E}_{\kappa}\frac{\rho\mu_1(\kappa)^2}{1+V(\kappa)}\\
\hat{q}_1=\frac{\alpha}{\beta}\mathbb{E}_{\kappa,y}\mu_1(\kappa)^2\frac{b(\kappa,y)^2+ \rho q(\kappa)-\mu_1(\kappa)^2\psi^2}{\left(1+V(\kappa)\right)^2}\\
\hat{V}_2=\frac{\alpha}{\beta}\mathbb{E}_{\kappa}\frac{\rho\mu_2(\kappa)^2}{1+V(\kappa)}\\
\hat{q}_2=\frac{\alpha}{\beta} \mathbb{E}_{\kappa,y}\mu_2(\kappa)^2\frac{b(\kappa,y)^2+ \rho q(\kappa)-\mu_1(\kappa)^2\psi^2}{\left(1+V(\kappa)\right)^2}\\
\hat{\zeta}=\frac{\alpha}{\sqrt{\beta}}\mathbb{E}_{\kappa,y}\kappa\mu_1(\kappa)\frac{b(\kappa,y)}{1+V(\kappa)}\\
\hat{\psi}=\frac{\alpha}{\sqrt{\beta}}\mathbb{E}_{\kappa,y}\frac{y\mu_1(\kappa)b(\kappa,y)+\psi\mu_1(\kappa)^2}{1+V(\kappa)}
\end{cases}
\end{align}
and
\begin{align}
    \rho^2=1-\gamma^2.
\end{align}
We introduced the shorthands
\begin{align}
    &b(y,\kappa)\equiv \sigma_\star(\gamma\kappa +\sqrt{1-\gamma^2} y)-\mu_0(\kappa)m -\kappa \mu_1(\kappa)\zeta-\mu_1(\kappa)\psi y\\
    &V(\kappa)\equiv \mu_1(\kappa)^2V_1+\mu_2(\kappa)^2V_2\\
    &q(\kappa)\equiv \mu_1(\kappa)^2q_1+\mu_2(\kappa)^2q_2,\\
    &I(\hat{V}_1,\hat{V}_2)=\int d\nu(\varrho,\tau,\pi)
    \frac{\tau^2\varrho }{\hat{V}_1\varrho+\hat{V}_2+\lambda }
\end{align}
In \eqref{eq:main:SP}, the expectations over $\kappa, y$ bear over standard Gaussian variables. We remind that the quantities $r,c,\gamma$ are characterized in Result \ref{res:GD_mapping}. Finally, the test error \eqref{eq:mse} admits the sharp characterization 
\begin{align}
\label{eq:test_error_replica}
    \epsilon_g=\mathbb{E}_{\kappa, y}\Bigg[&\Bigg(
    \sigma_\star(\gamma\kappa +\sqrt{1-\gamma^2} y)-\mu_0(\kappa)m-\mu_1(\kappa)\kappa\zeta-\frac{\mu_1(\kappa)\psi}{\sqrt{\rho}} y
    \Bigg)^2 +q(\kappa)-\frac{\mu_1(\kappa)^2\psi^2}{\rho} \Bigg]
\end{align}
where expectations bear over standard Gaussian variables.
\end{result}
Result \ref{res:asymptotics} thus rephrases the high-dimensional learning problem \eqref{eq:ERM_a} in terms of a finite set of scalar summary statistics which can be efficiently evaluated, yielding excellent agreement with finite $d$ numerical simulations, see Figs. \ref{fig:GD}-\ref{fig:VaryR}. We emphasize that while the solid lines indicate the theoretical characterization of the test error achieved by the sRF, the numerical experiments were performed for the \textit{original two-layer network setting}. The good agreement thus further supports that sRFs constitute good models of the latter. While we state Result \ref{res:asymptotics} for the square loss and $\ell_2$ regularization for clarity, we provide a sharp characterization for any generic convex loss $\ell$ in Appendix \ref{sec:app:replica}.\\ 

As we discuss in further detail in Appendix \ref{sec:app:replica}, the derivation of the result using the replica method is sizeably more involved than that for unspiked RF models, as reported in \cite{gerace_generalisation_2020}. Unlike the classic Gaussian equivalence which holds for vanilla RFs, Result \ref{res:asymptotics} is conditioned on the variable $\kappa$, which needs to be handled specifically in the computation, borrowing ideas from the superstatistical approach of \cite{adomaityte2023high}, see also the discussion between equations (94) and (95) in Appendix \ref{sec:app:replica}. Another difficulty lies in the treatment of the spike, absent in classic random features, which necessitates the tracking and non-standard asymptotic control of more summary statistics. \\

Finally, while we have used the (non-rigorous) replica method to derive these equations, an interesting avenue of future research is to provide a rigorous probabilistic proof. A possible way to address the problem is to 
 apply Gordon’s Gaussian min-max inequalities \cite{gordon1988milman,thrampoulidis2014gaussian,stojnic2013meshes,stojnic2013upper}), -- and extend the approach of \cite{mignacco2020role} for binary Gaussian mixtures to the infinite mixture the equivalent that the feature distribution \eqref{eq:phig} is formally equivalent to--; and generalize the approach used for pure random features \cite{hu2022universality,loureiro2021learning}. This is, however, beyond the scope of this manuscript.\\

Finally, our results also allow to sharply characterize the Stieltjes transform (and thus the spectrum) of the empirical features covariance $\sfrac{1}{n}\varphi(X)^T\varphi(X)$, denoting $X\in\mathbb{R}^{n\times d}$ the matrix with rows $\{\vec{x}^\mu\}_{\mu=1}^n$. This analysis is presented in detail in Appendix \ref{sec:app:Stieltjes}, and reveals an intimate connection between the Stieltjes transform and the summary statistic $V_2$ in \eqref{eq:SP}. This result complements the spectral analysis of \cite{moniri2023theory} for step sizes $\eta=\Theta_d(d^s)$ for $\sfrac{1}{2}< s <1$. In a close setting, \cite{wang2022spectral} cover the $s=\sfrac{1}{2}$ case, analyzing the conjugate kernel of deep network features for spiked data. Finally, the works of \cite{guionnet2023spectral, feldman2023spectral} analyze the spectrum of related non-linear spiked Wigner matrices.

%% file: sections/Derivation.tex
\section{Discussion of Main Results}
While the self-consistent equations in Result \ref{res:asymptotics} might appear cumbersome, they offer valuable insight into the mechanism behind feature learning in two-layer neural networks trained with gradient descent. In this section, we discuss and highlight some of these insights.
\begin{figure}[t]
    \centering
    \includegraphics[scale=0.66]{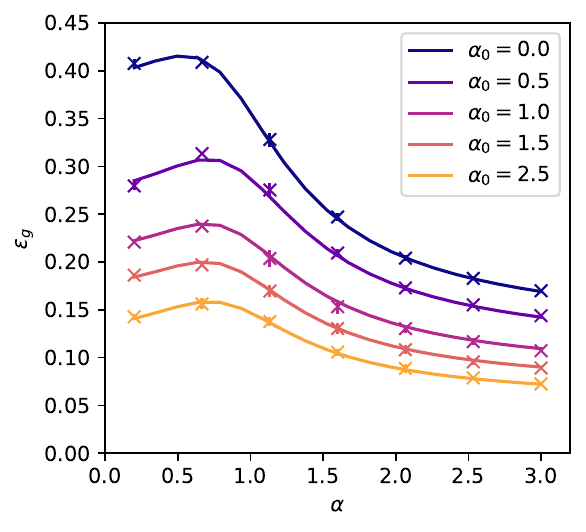}
    \includegraphics[scale=0.66]{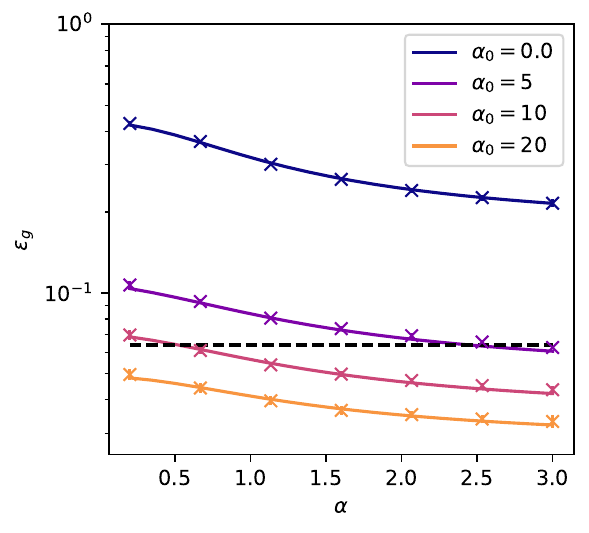}
    \caption{Crosses: numerical evaluation of the test error achieved by a two-layer network whose first layer has been trained following the protocol detailed in section~\ref{sec:motivation}, with learning rate $\Tilde{\eta}=1$, activation $\sigma=\tanh$, and readout regularization (\textbf{left}) $\lambda=0.01$ (\textbf{right}) $\lambda=0.1$ . The target is a single index model with (\textbf{left}) tanh  (\textbf{right}) sine activation. Numerical experiments were performed in $d=2000$. All points were averaged over $5$ instances. Different colours represent different initial sample complexities $\alpha_0=\sfrac{n_0}{d}$ used for the first gradient step. Solid lines: theoretical characterization of Result \ref{res:asymptotics} for the equivalent sRFs. The dashed black line represents the lowest achievable MSE for kernel/linear methods, namely $h^\star_2-(h^\star_1)^2$ \cite{ba2022high}. }
    \label{fig:GD}
\end{figure}

\subsection{Spiked Random Features vs Random Features}
\begin{figure}
    \centering
    \includegraphics[scale=0.66]{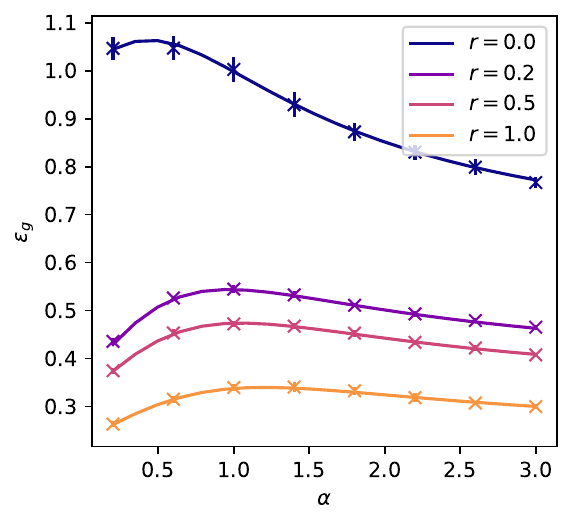}
    \includegraphics[scale=0.66]{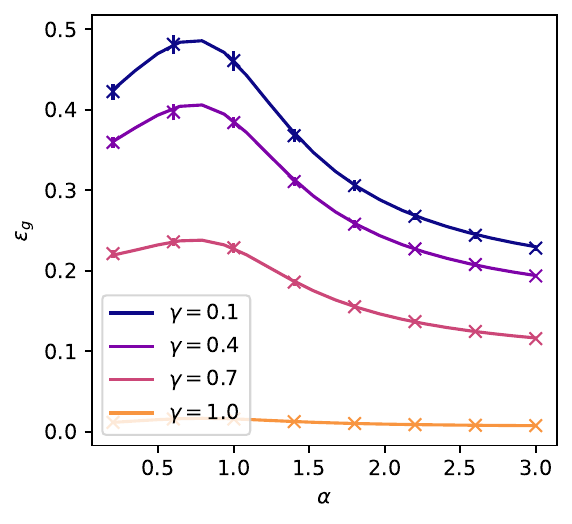}
    \caption{Test error for a sRF with (\textbf{left}) $\sigma=\sin$ (\textbf{right}) $\sigma=\tanh$ activation, learning from a single-index model (\textbf{left}) $\mathrm{sign}(\sfrac{\vec \theta^\top \vec x}{\sqrt{d}})$ (\textbf{right}) $\tanh(\sfrac{\vec \theta^\top \vec x}{\sqrt{d}})$, with regularization $\lambda=0.1$. Solid lines: theoretical characterization of Result \ref{res:asymptotics}. Crosses: numerical simulations in dimensions $d=p=2000$. Each point is averaged over $10$ instances of the problem. Different colours correspond to different spike strengths $r$ \eqref{eq:W}, with $r=0$ corresponding to the vanilla RF model.}
    \label{fig:VaryR}
\end{figure}

The asymptotic characterization \ref{res:asymptotics} encompasses, as a special case, usual RFs (when setting the spike strength $r$ to zero). More precisely, for zero spike strength $r=0$ in \eqref{eq:W}, sRFs coincide with RFs, as the coefficients \eqref{eq:main:mus} lose their $\kappa$ dependence, reducing to usual Hermite coefficients. The equivalent feature map $\vec \varphi^g$ then reduces to the Gaussian equivalent feature map employed in e.g. \cite{goldt2020modeling, hu2022universality, schroder2023deterministic}. Importantly, while the equivalent feature map for unspiked RFs is \textit{linear} in the input $\vec x$, the equivalent feature map $\vec \varphi^g$ \eqref{eq:phig} of Result \ref{conj:main:gaussian_equivalence} is \textit{non-linear} in the component $\kappa$ of the input $\vec x$.\looseness=-1\\

sRF models, which provide good models for two-layer networks after a single large gradient step, offer an ideal playground to test the intertwined influence of the spike/target correlation and the test performance of the model. Fig.\,\ref{fig:VaryR} presents the learning curves of (s)RFs for varying spike strengths $r$. As is intuitive, larger spikes allow the sRF to more easily express non-linear function in the direction of the spike, and thus lead to relatively smaller test errors. Furthermore, note that even rather small spike strengths $r=0.2$ already yield test errors which are sizably lower than vanilla (unspiked) RFs ($r=0$), hinting at the qualitative difference between sRF and RF models discussed in section \ref{sec:motivation} and Fig.\,\ref{fig:NL}. The plot shows a compelling agreement between the theoretical predictions (continuous line) and numerical simulations.  Fig.\,\ref{fig:VaryR} represents the theoretical closed-form expression for the test error \eqref{eq:test_error_replica} for different values of the spike/target alignment $\gamma$, for a single-index target $\mathrm{sign}(\vec \theta^\top \vec x)$, with good agreement with numerical experiments. Higher alignments $\gamma$ lead to overall lower test errors escaping the linear curse of Gaussian models. 

%%%%%%%%%%%%%%%%%%%%%%%%%%%%%%%%%%%%%%%%%%%%%%%%%
\subsection{Beating kernels in a single step}
\label{sec:bounds}
%%%%%%%%%%%%%%%%%%%%%%%%%%%%%%%%%%%%%%%%%%%%%%%%%
A key parameter in our formulas is $\gamma = \sfrac{\vec{v}^{\top}\vec{\theta}}{d} \in [0,1]$, the correlation between the effective gradient spike and the target weights. From its asymptotic expression eq. \ref{eq:GD_sRF} in Result \ref{res:GD_mapping}, this is an increasing function of $\alpha_{0}=\sfrac{n_{0}}{d}$, the number of samples in the first step. As shown in Fig.~\ref{fig:GD} for $\sigma=\sigma_\star=\tanh$, larger sample complexities in the representation learning step $\alpha_0$ allow for better feature learning when implementing a gradient step on the first layer, enabling a lower test error after the readout layer is retrained. As expected, the lowest error is achieved when $\alpha_{0}\to\infty$, in which case the spike $\vec{v}$ is perfectly aligned with the target weights $\vec{\theta}$ and $\gamma=1$, as can be seen from \eqref{eq:GD_sRF}.\\

Fig.\,\ref{fig:GD} presents similar curves for another target activation $\sigma^\star=\sin$, including the test error achieved by the network at initialization ($\alpha_0=0$), which corresponds to a usual RF model. The latter is well above the lowest Mean Squared Error (MSE) achievable by a linear estimator (plotted as a dashed black line), namely the projection of the teacher function on Hermite polynomials $\lVert \mathbb{P}_{>1}f_\star \lVert_2^2$. This performance corresponds to the 
kernel one when the number of samples scales proportionally with the input dimension \cite{Ghorbani2020}; using the notations of Result~\ref{res:GD_mapping} the best linear MSE is readily written as $h^\star_2-(h^\star_1)^2$. Note that this value also lower-bounds the MSE achieved by the NTK, see Proposition $1$ of \cite{ba2022high}. In sharp contrast, networks with trained first layers ($\alpha_0>0$) can learn non-linear functions of the inputs and outperform this baseline. 

\subsection{What can be learned with a single step?}
\label{sec:what}
While training a two-layer network with large gradient steps allows it to escape the linear limitations of, e.g. RF models, it generically only learns the target up to small test error, yet not perfectly. Further insight can be extracted from the theoretical characterization of the corresponding sRF. In fact, a closer examination of the sharp asymptotic expression \eqref{eq:test_error_replica} for $\epsilon_g$ in Result \ref{res:asymptotics} reveals that even for large initial batches $\alpha_0\to\infty$ (and thus perfect spike/target alignments $\gamma=1$), at fixed learning rate strength $\Tilde{\eta}$ and sample complexity $\alpha>0$, the test error optimized over the regularization $\lambda$ is upper bounded as
\begin{align}
\label{eq:upper_bound}
    \underset{\lambda\ge0}{\inf}\epsilon_g\le \underset{b_1}{\inf}~\mathbb{E}_{\kappa}\left[
    \sigma_\star(\kappa)-b_1\mu_0(\kappa)
    \right]^2,
\end{align}
and lower bounded by:
\begin{align}
\label{eq:lower_bound}
    \underset{\lambda\ge 0}{\inf}~\epsilon_g\!\ge\! \underset{b_1, b_2}{\inf}~\mathbb{E}_{\kappa}\left[
    \sigma_\star(\kappa)-b_1\mu_0(\kappa)-b_2\mu_1(\kappa)\kappa
    \right]^2.
\end{align}
\begin{wrapfigure}{r}{0.5\textwidth}
    \centering
    \includegraphics[scale=0.66]{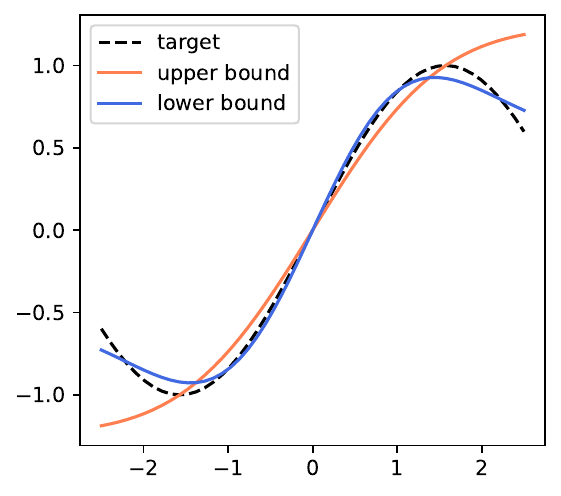}
    \caption{($c=1, r=0.9 $ and $ \gamma=1$) Illustration of the functions  realizing the upper bound \eqref{eq:upper_bound} (orange) and lower bound \eqref{eq:lower_bound} (blue), for $\sigma=\tanh$, for a target $\sigma_\star=\sin$ (dashed black). }
    \label{fig:bounds}
\end{wrapfigure}
The upper bound \eqref{eq:upper_bound} is the equivalent of Lemma 6 of \cite{ba2022high} in our case of uniform readout initialization $\vec a^{(0)}=\sfrac{\vec 1_{p}}{\sqrt{p}}$, and is achieved for $\lambda \to\infty$, as we show in App. \ref{app:bounds}.  The lower bound \eqref{eq:lower_bound}, however, shows that the test error cannot be lower than the Gaussian-weighted $L^2$ distance between the target function $\sigma_\star$ and $\mathrm{span}(\mu_0, \tilde{\mu}_1)$, where $\tilde{\mu}_1(\kappa)\equiv \kappa \mu_1(\kappa)$, and the best approximation would be reached for the projection of the target thereon. Fig. \ref{fig:bounds} illustrates what functions realize the upper  \eqref{eq:upper_bound} and lower bounds \eqref{eq:lower_bound}, and how they compare with the target $\sigma_{\star}$. Finally, note that in the vanilla RF limit $r=0$, the functions $\mu_{0}(\kappa),\mu_{1}(\kappa)$ reduce to constants independent of $\kappa$, constraining the class of functions that can be learned to that of linear functions.\\

Finally, let us mention that while this discussion was made at fixed learning rate $\tilde{\eta}$ for clarity, the latter is in practice a tunable hyper-parameter, and the functions $\mu_0,\mu_1$ depend thereupon via the spike strength $r$ \eqref{eq:main:mus}.  One can thus refine -- and lower-- the bounds \eqref{eq:upper_bound} and \eqref{eq:lower_bound} over the functions $\mu_0,\mu_1$, which can take values in the realizable set $\mathcal{M}=\{(\mu_0(r),\mu_1(r)\}_{r\ge 0}$ (emphasizing the dependence on $r$ via equation \eqref{eq:app:mus}) as $\Tilde{\eta}$ is varied: 
\begin{align}
\label{eq:bound_opteta}
    &\underset{\lambda\ge0,\tilde{\eta}\ge 0}{\inf}\epsilon_g\le \underset{b_1\in\mathbb{R},\nu_0\in\mathcal{M}}{\inf}\mathbb{E}_{\kappa}\left[
    \sigma_\star(\kappa)-b_1\nu_0(\kappa)
    \right]^2,\\
    &\underset{\lambda\ge 0,\tilde{\eta}\ge 0}{\inf}\epsilon_g\ge\!\!\!\!\!\!\!\!\! \underset{\begin{array}{cc}
          {\scriptstyle b_1,b_2\in\mathbb{R}}, \\
        {\scriptstyle(\nu_0,\nu_1)\in\mathcal{M}}
    \end{array}}{\inf}\!\!\!\!\!\!\!\!\!\mathbb{E}_{\kappa}\left[
    \sigma_\star(\kappa)-b_1\nu_0(\kappa)-b_2\nu_1(\kappa)\kappa
    \right]^2.\notag
\end{align}
In other words, by tuning the learning rate $\Tilde{\eta}$ -- and thus the spike strength $r$--, one gains the freedom to choose the "best" subspace $\mathrm{span}(\mu_0(r),\mu_1(r))$, i.e. the one allowing to approximate the target $\sigma_\star$ best. Finally, as discussed in \cite{ba2022high}, observe that for $\sigma=\sigma_\star=\mathrm{erf}$, the upper bound in \eqref{eq:bound_opteta} is zero provided one tunes the learning rate to $\tilde{\eta}=\sfrac{\sqrt{3\beta}}{h_1}$, and perfect learning is therefore achievable. Let us emphasize that this discussion formally holds for the equivalent sRF model; we however expect it to hold, at least qualitatively, also for the original two-layer networks.

\subsection{More variability means better feature learning}

The discussion of subsection \ref{sec:what} thus affords an insightful perspective on the learning of sRF models in terms of approximating the target activation $\sigma_\star$ in a two-dimensional functional space.
Interestingly, introducing variability in the readout layer at initialization leads to an even richer functional basis, and hence greater expressivity of the network. As we further discuss in Appendix \ref{app:non_uniform}, when $\vec{a}^{(0)}$ is no longer proportional to $\vec 1_p$, but is rather initialized from a distribution over a finite vocabulary $V$ of size $|V|>1$--e.g. $V = \{-1,0,+1\}$-- the equivalent feature map of the associated sRF takes the form:\looseness=-1
\begin{align}
\label{eq:non-unif-phi}
    \vec{\varphi}^g(\vec{x})\!\!= \!\!\begin{pmatrix}
    \scriptstyle
    \scriptstyle\mu_0(u_1\kappa)\\
    \scriptstyle\mu_0(u_2\kappa)\\
    \vdots\\
    \scriptstyle\mu_0(u_p\kappa)
    \end{pmatrix}
    + \begin{pmatrix}
    \scriptstyle\mu_1(u_1\kappa)\\
    \scriptstyle\mu_1(u_2\kappa)\\
    \vdots\\
    \scriptstyle\mu_1(u_p\kappa)
    \end{pmatrix} \odot \W \vec{x}+ \begin{pmatrix}
    \scriptstyle\mu_2(u_1\kappa) \\
    \scriptstyle\mu_2(u_2\kappa)\\
    \vdots\\
    \scriptstyle\mu_2(u_p\kappa)
    \end{pmatrix} \odot \vec{\xi}
\end{align}
where $\odot$ denotes element-wise multiplication, $\vec{\xi}\sim\mathcal{N}(0, \mathbb{I}_p)$, and $\vec u\not\propto \vec1_p$  has entries which can now take a finite number of different values. The coefficients of the equivalent map \eqref{eq:non-unif-phi} are thus neuron-dependent and thus afford a richer functional basis $\{\mu_0(\omega \cdot), \tilde{\mu}_1(\omega \cdot)\}_{\omega\in V}$, thereby allowing the network to express a larger class of functions. As a matter of fact, the functional space spanned by these functions is generically of dimension $2|V|$ for non-uniform readout initializations $\vec a^{(0)}$, compared to just $2$ in the uniform readout case. A sharp asymptotic characterization of the test error for the case of non-uniform readout initialization with finite vocabulary is further provided in Appendix \ref{app:non_uniform}.\\

We briefly discuss the limiting case of interest $\lambda, \alpha_0, \Tilde{\eta} \rightarrow \infty$, for which the equivalent feature map $\vec{\varphi}^g(\vec{x})$ \eqref{eq:non-unif-phi} reduces to its first term $(\mu_0(u_i\kappa))_{i=1}^p$. Further, observe that $\mu_0(u_i\kappa)$ can be viewed as a one-dimensional neuron acting on the one-dimensional input variable $\kappa$ with a random weight $u_i$. Standard results on approximation errors for random feature mappings of finite-dimensional inputs \cite{rahimi2007random} imply that a large class of functions can be approximated from the network features $\vec{\varphi}^g(\vec{x})$, provided the vocabulary size $|V|$ is large enough. Similar random features approximations have been leveraged in \cite{ba2022high,ba2023learning,damian2022neural}. The equivalent feature map \eqref{eq:non-unif-phi} provides an %particularly 
intuitive picture on how such random feature mappings of low-dimensional inputs can naturally emerge in the setting of the learning of a two-layer network.\looseness=-1\\

Finally, note that the bounds \eqref{eq:upper_bound} and \eqref{eq:lower_bound} can be readily generalised to non-uniform readout initializations, provided one replaces in the discussion of subsection \ref{sec:what} the two-dimensional functional basis $\mathrm{span}(\mu_0, \tilde{\mu}_1)$ in the case of uniform initialization by the richer functional space $\mathrm{span}(\{\mu_0(\omega \cdot), \tilde{\mu}_1(\omega \cdot)\}_{\omega\in V})$ for non-uniform initialization. For instance, the lower bound \eqref{eq:lower_bound} thus involves in the non-uniform case the distance between the target $\sigma_\star$ and $\mathrm{span}(\{\mu_0(\omega \cdot), \tilde{\mu}_1(\omega \cdot)\}_{\omega\in V})$.

%%%%%%%%%%%%%%%%%%%%%%%%
\section*{Conclusion}
%%%%%%%%%%%%%%%%%%%%%%%%
We provided a tight asymptotic description of the learning of spiked random features models, which we show provides a quantitative approximation for a two-layer neural network after training its first layer with a large, single, gradient step, in the limit where the number of samples, the hidden layer width, and the input dimension are proportionally large. Our results sharply characterize the corresponding feature map, and how it achieves a test error which non-perturbatively improves over the kernel regime. Crucially, the sRF can efficiently approximate non-linear functions --beyond polynomials-- in the direction of the gradient -- sizeably improving upon RFs (modeling the network at initialization), which can only express linear functions. We further discuss bounds for the test error and which functions are learnable after a single gradient step. Finally, extensive numerical support is provided to illustrate our findings. \\

We believe the present work opens exciting research avenues, paving the way towards a tight theoretical understanding of feature learning in gradient-trained networks. Prominent among these research directions is the extension of our results to readout initialization with generic (not necessarily finite) support.  to a finite number of gradient steps, and ultimately fully trained networks.

%% file: sections/acknowledgements.tex
\section{Acknowledgements}
We thank Denny Wu, Gabriel Arpino, Weiyu Li, Marco Mondelli and Theodor Misiakiewicz for discussion during the course of this project. Early discussions on this project started as an open problem posed by BL during the 2022 Les Houches Summer School on \emph{Statistical Physics and Machine Learning} organized by LZ and FK. We acknowledge funding from the Swiss National Science Foundation grant SNFS OperaGOST  (grant number $200390$), and SMArtNet (grant number $212049$), and the Choose France - CNRS AI Rising Talents program.

%% file: sections/appendix/mapping.tex
\section{Derivation of Result \ref{res:GD_mapping}}
\label{app:GD}
This Appendix details the derivation of Result \ref{res:GD_mapping}, which provides tight asymptotic formulae \eqref{eq:GD_sRF} for the parameters of the sRF
associated to a two-layer neural network trained with a large gradient step.

\subsection{First layer weights after one gradient step --}

After one large gradient step, the weights $W^{(1)}$ can be decomposed as a bulk plus a spike in the following way:

\begin{align}
    \label{eq:spikebulk}
    W^{(1)}=W^{(0)}+\Delta+\frac{\check{\vec u}\check{\vec v}^\top}{\sqrt{p}},
\end{align}
where 
\begin{align}
    \label{eq:u_v}
    &\check{\vec u}=\eta h_1 \vec a^{(0)},\\
    & \check{\vec v}=\frac{1}{n_0}\sum\limits_{\mu=1}^{n_0} f_\star(\vec x^\mu)\vec x^\mu,
    \end{align}
and
\begin{align}
\label{eq:Delta}
\Delta=W^{(0)}+\frac{\eta }{n_0\sqrt{p}}\mathrm{diag}(\vec a^{(0)})\left(
    \sum\limits_{\mu=1}^{n_0}\check{\sigma}(W^{(0)}\vec x^\mu)f_\star(\vec x^\mu)\vec x^\mu
    \right).
\end{align}
We noted $h_1$ the first Hermite coefficient of the student activation $\sigma$ and $\check{\sigma}(\cdot)=\sigma^\prime(\cdot)-h_1$. As discussed in \cite{ba2022high, dandi2023universality}, the bulk $\Delta$ behaves as an effective noise which rescales the initial weights $W^{(0)}$; we refer to Sec.~{\bf{B.6}} of \cite{dandi2023twolayer} for additional rigorous characterization of the bulk $\Delta$. To connect to the convention used below \eqref{eq:W} of normalized $\vec u= \vec 1_p$, $\lVert \vec v\lVert =\sqrt{d}$, and a single bulk $\W$, one thus needs to assess the asymptotic limits of $r=\sfrac{\lVert \check{\vec u}\lVert\lVert \check{\vec v}\lVert}{\sqrt{pd}}$, $c=\sfrac{1}{\sqrt{d}}\lVert W^{(0)}+\Delta\lVert$ and $\gamma=\sfrac{\check{\vec v}^\top \vec \theta}{\lVert \check{\vec v}\lVert}$. This program is carried out in the following subsection.

\subsection{Spike scale}
We first focus on the scale of the spike $r$. We remind here the convention to assume normalized spike vectors, i.e. $\lVert \vec u\lVert =\lVert \vec v\lVert=\sqrt{d}$. Comparing to \eqref{eq:u_v}, it follows that
\begin{align}
    r=\frac{1}{\sqrt{pd}}\left\lVert
    \eta h_1 a^{(0)}
    \right\lVert
    \times
    \left\lVert
\frac{1}{n_0}\sum\limits_{\mu=1}^{n_0} f_\star(\vec x^\mu) \vec x^\mu
    \right\lVert.
\end{align}
Remembering that $\vec a^{(0)}=\sfrac{\vec 1_p}{\sqrt{p}}$, it follows that
\begin{align}
    \left\lVert
    \eta h_1 \vec a^{(0)}
    \right\lVert=d\Tilde{\eta} h_1.
\end{align}
We now turn to the second term. As a mean of independent vectors, it concentrates in probability. We thus compute the expectation of the squared norm:
\begin{align}
\mathbb{E}_{\mathcal{D}_0} \left\lVert
\frac{1}{n_0}\sum\limits_{\mu=1}^{n_0} f_\star(\vec x^\mu)\vec x^\mu
    \right\lVert^2&=\frac{1}{n_0}\sum\limits_{i=1}^d\mathbb{E}_{\vec x}[f_\star(\vec x)^2 x_i^2]+\frac{(n_0-1)}{n_0}\sum\limits_{i=1}^d\mathbb{E}_x[f_\star(\vec x) x_i]^2
\end{align}
Note that $\sfrac{\vec \theta^\top \vec x}{\sqrt{d}}$ and $x_i$ have asymptotically vanishing correlation. To leading order, therefore
\begin{align}
    & \mathbb{E}_{\vec x}[f_\star(\vec x)^2 x_i^2]=\mathbb{E}_z[\sigma_\star(z)^2]+O_d(\sfrac{1}{\sqrt{d}})=h_2^\star\\
    &\mathbb{E}_{\vec x}[f_\star(\vec x) x_i]=\mathbb{E}_z[\sigma_\star^\prime(z)]\frac{\theta_i}{\sqrt{d}}= h_1^\star \frac{\theta_i}{\sqrt{d}}
\end{align}
Therefore
\begin{align}
    \mathbb{E}_{\mathcal{D}_0} \left\lVert
\frac{1}{n_0}\sum\limits_{\mu=1}^{n_0} f_\star(\vec x^\mu) \vec x^\mu
    \right\lVert^2&=\frac{d h_2^\star}{n_0}+\frac{(n_0-1)h_1^{\star2}}{n_0}=\frac{h_2^\star}{\alpha_0}+h_1^{\star 2}.
\end{align}

Thus 
\begin{align}
    r=\frac{\Tilde{\eta}h_1}{\sqrt{\beta}}\left(\frac{h_2^\star}{\alpha_0}+h_1^{\star 2}\right)^{\sfrac{1}{2}}.
\end{align}

\subsection{Spike/target alignment}
We now turn to the spike/target alignment, defined as the overlap between the (normalized) target weights $\theta$ and spike $v$.
\begin{align}
    \gamma=\frac{\vec \theta^\top \vec v\sqrt{d}}{d\lVert \vec v\lVert}=\frac{\frac{1}{n_0}\sum\limits_{\mu=1}^{n_0} \sigma_\star(z^\mu)z^\mu}{\left\lVert
\frac{1}{n_0}\sum\limits_{\mu=1}^{n_0} f_\star(x^\mu)x^\mu
    \right\lVert}.
\end{align}
We noted $z^\mu\equiv\sfrac{\vec \theta^\top \vec x^\mu}{\sqrt{d}}$. The numerator converges by the law of large numbers to $h^\star_1$, while the numerator was computed in the previous subsection. All in all, we reach
\begin{align}
    \gamma=\frac{h_1^\star}{\left(\frac{h_2^\star}{\alpha_0}+h_1^{\star 2}\right)^{\sfrac{1}{2}}}
\end{align}

\subsection{Bulk norm}
By symmetry, all rows of the bulk $\Delta$ \eqref{eq:Delta} asymptotically share the same norm $c_\Delta$. The norm of a row $w$ of $\W=W^{(0)}+\Delta$ thus has asymptotic norm 
\begin{align}
    \lVert w_i\lVert =\sqrt{1+c_\Delta^2+2c_{W\Delta}}
\end{align}
where
\begin{align}
    &c_\Delta=\left\lVert\frac{\eta }{n_0p}\left(
    \sum\limits_{\mu=1}^{n_0}\check{\sigma}(w^{(0)\top}x^\mu)f_\star(x^\mu)x^\mu
    \right)\right\lVert\\
    & c_{W\Delta}=\frac{\eta }{n_0p}\left(
    \sum\limits_{\mu=1}^{n_0}\check{\sigma}(w^{(0)\top}  x^\mu)f_\star(x^\mu)w^{(0)\top} x^\mu
    \right),
\end{align}
where with a slight abuse of notation we dropped the row index $i$, since all rows are equivalent. 
\begin{align}
    \mathbb{E}_{\mathcal{D}_0}   [c_\Delta^2]=\frac{\eta^2}{n_0p^2} \sum\limits_{i=1}^d\mathbb{E}_x[\check{\sigma}(w^{(0)\top}x)^2\sigma_\star(\sfrac{\theta^{\top}x}{\sqrt{d}})^2x_i^2]+\frac{(n_0-1)\eta^2}{n_0p^2} \sum\limits_{i=1}^d\mathbb{E}_x[\check{\sigma}(w^{(0)\top}x)\sigma_\star(\sfrac{\theta^{\top}x}{\sqrt{d}}) x_i]^2
\end{align}
Because $w^{(0)\top}x,\sfrac{\theta^{\top}x}{\sqrt{d}},x_i $ all have asymptotically vanishing $\Theta_d(\sfrac{1}{\sqrt{d}})$ correlations, the first summand reads to leading order
\begin{align}
    \mathbb{E}_x[\check{\sigma}(w^{(0)\top}x)^2\sigma_\star(\sfrac{\theta^{\top}x}{\sqrt{d}})^2x_i^2]=\check{h}_1^2 h_2^\star
\end{align}
The second summand reads, using Stein's lemma
\begin{align}
    \mathbb{E}_x[\check{\sigma}(w^{(0)\top}x)\sigma_\star(\sfrac{\theta^{\top}x}{\sqrt{d}}) x_i]&=w^{(0)}_i\mathbb{E}_x[\check{\sigma}^\prime(w^{(0)\top}x)\sigma_\star(\sfrac{\theta^{\top}x}{\sqrt{d}}) ]+\frac{\theta_i}{\sqrt{d}}\mathbb{E}_x[\check{\sigma}(w^{(0)\top}x)\sigma_\star^\prime(\sfrac{\theta^{\top}x}{\sqrt{d}})]\notag\\
    &=\Theta_d(\sfrac{1}{d})
\end{align}
We used
\begin{align}
    \mathbb{E}_z[\check{\sigma}(z)]=\mathbb{E}_z[\check{\sigma}^\prime(z)]=0
\end{align}
respectively by definition of $\check{\sigma}$ and because $\sigma$ is odd. Thus 
\begin{align}
    \mathbb{E}_{\mathcal{D}_0}   [c_\Delta^2]=\frac{\eta^2 \check{h}_1^2 h_2d}{n_0p^2} +\Theta_d(\sfrac{1}{d})=\frac{\Tilde{\eta}^2 \check{h}_1^2 h_2^\star}{\alpha_0 \beta^2} +o_d(1).
\end{align}
The cross-term $c_{\Delta W}$ can similarly be computed as
\begin{align}
    c_{W\Delta}=\frac{\eta }{p}\mathbb{E}_x[\check{\sigma}(w^{(0)\top}  x)\sigma_\star(\sfrac{\theta^{\top}x}{\sqrt{d}})w^{(0)\top} x]
\end{align}
Again because all variables in the expectation are weakly correlated, and because the average of $\check{\sigma}$ is zero, we have that $c_{W\Delta}$ is vanishing to leading order.
\begin{align}
    c_{W\Delta}=o_d(1)
\end{align}
Finally, we thus have
\begin{align}
    c=1+\frac{\Tilde{\eta}^2 \check{h}_1^2 h_2^\star}{\alpha_0 \beta^2}
\end{align}

\subsection{Summary of the mapping}
In summary, we have derived the asymptotic formulae for the equivalent sRF:
\begin{align}
    &c=1+\frac{\Tilde{\eta}^2 \check{h}_1^2 h_2^\star}{\alpha_0 \beta^2}\\
    &r=\frac{\Tilde{\eta}h_1}{\sqrt{\beta}}\left(\frac{h_2^\star}{\alpha_0}+h_1^{\star 2}\right)^{\sfrac{1}{2}},\\
    &\gamma=\frac{h_1^\star}{\left(\frac{h_2^\star}{\alpha_0}+h_1^{\star 2}\right)^{\sfrac{1}{2}}},
\end{align}
which recovers equation \eqref{eq:GD_sRF} of Result \ref{res:GD_mapping}.

%% file: sections/appendix/cGET.tex
\section{Derivation of Result \ref{conj:main:gaussian_equivalence}}
\label{app:CGET}

In this section we detail the heuristic derivation of Result \ref{conj:main:gaussian_equivalence}, which shows the asymptotic equivalence of the sRF feature map \eqref{eq:phi} with the conditional feature map \eqref{eq:phig}. This mapping is at the core of the exact asymptotic description detailed in the next appendix. 
\\We first remark that, since the data is Gaussian, $x\sim \mathcal{N}(0,\mathbb{I}_d)$, we can decompose it as 
\begin{align}
    x=\kappa +x^\perp,
\end{align}
where $\kappa\equiv\sfrac{v^\top x}{\sqrt{d}}\sim \mathcal{N}(0,1)$ and we noted $x^\perp\equiv\Pi^\perp x$ the component orthogonal to the spike, with Gaussian statistics $x^\perp\sim\mathcal{N}(0,\Pi^\perp).$ The sRF feature map \eqref{eq:phi} thus reads
\begin{align}    \varphi(x)=\sigma\left(\kappa 1_p + \kappa \frac{\W v}{\sqrt{d}}+\W x^\perp \right)
\end{align}
Conditioning on the projection $\kappa$ along the spike, this corresponds to a RF feature map with random biases $\kappa 1_p +\kappa\sfrac{\W v}{\sqrt{d}}$. One can compute the population mean and covariance of $\varphi_x$ along the exact same lines as e.g. \cite{cui2023bayes}, leading to 
\begin{align}
    &\mathbb{E}_x[\varphi(x)_i]=\Tilde{\kappa}^0_i.\\
    &\mathrm{Cov}_x[\varphi(x)_i,\varphi(x)_j]= \Tilde{\kappa}^1_i\Tilde{\kappa}^1_j (\W)_i^\top \Pi^\perp (\W)_j+\delta_{ij} (\kappa^*_i)^2.
\end{align}
where
\begin{align}
\label{eq:defkappas}
    &\Tilde{\kappa}^0_i\equiv \mathbb{E}_z[\sigma(\kappa+\sfrac{(\W)_i^\top v}{\sqrt{d}}+z)]\\
    &\Tilde{\kappa}^1_i\equiv \mathbb{E}_z[z\sigma(\kappa+\sfrac{(\W)_i^\top v}{\sqrt{d}}+z)]\\
    &\Tilde{\kappa}^*_i=\sqrt{\mathbb{E}_z[\sigma(\kappa+\sfrac{(\W)_i^\top v}{\sqrt{d}}+z)^2]-(\Tilde{\kappa}^1_i)^2-(\Tilde{\kappa}^0_i)^2}
\end{align}
In other words, the features $\varphi(x)$ \textit{share the same second order statistics} as the noisy features
\begin{align}
    \varphi^g(x)_i\equiv \Tilde{\kappa}^0_i+\Tilde{\kappa}^1_i (\W)_i^\top x^\perp +\Tilde{\kappa}^*_i \xi_i
\end{align}
where $\xi_i\sim\mathcal{N}(0,1)$ is a stochastic noise. Note that in the definition of the $\kappa^{0,1,*}_i$ coefficients \eqref{eq:defkappas}, $\sfrac{(\W)_i^\top v}{\sqrt{d}}=\Theta_d(\sfrac{1}{\sqrt{d}})$, so one can expand them as
\begin{align}
\label{eq:defkappas_expanded}
    &\Tilde{\kappa}^0_i\equiv \underbrace{\mathbb{E}_z[\sigma(\kappa+z)]}_{\mu_0(\kappa)}+\sfrac{(\W)_i^\top v}{\sqrt{d}} \mu_1(\kappa)\\
    &\Tilde{\kappa}^1_i\equiv \underbrace{\mathbb{E}_z[z\sigma(\kappa+z)]}_{\mu_1(\kappa)}+\sfrac{(\W)_i^\top v}{\sqrt{d}}
    \underbrace{\mathbb{E}_z[z\sigma^{\prime}(\kappa+z)}_{\equiv \mu_3(\kappa)}]
\\    &\Tilde{\kappa}^*_i=\underbrace{\sqrt{\mathbb{E}_z[\sigma(\kappa+z)^2]-(\kappa^1)^2-(\kappa^0)^2}}_{\mu_2(\kappa)}+O_d(\sfrac{1}{\sqrt{d}})
\end{align}
So
\begin{align}
    \varphi^g(x)_i &\equiv \mu_0(\kappa)+\mu_1(\kappa) (\W)_i^\top x+ \\ 
    &+\mu_3(\kappa)\times \frac{(\W)_i^\top v}{\sqrt{d}}\times(\W)_ix^\perp+\mu_2(\kappa) \xi_i+\xi_i\Theta_d(\sfrac{1}{\sqrt{d}})
\end{align}
Finally, note that the $\mu_3(\kappa)$ term is subleading compared to the second term, and should be asymptotically unconsequential for the learning, and can thus be safely neglected. By the same token, the last term is a subleading correction to the noise term and can be neglected. Therefore, one finally reaches
\begin{align}
    \varphi^g(x)\equiv \mu_0(\kappa)1_p+\mu_1(\kappa) \W^\top x+\mu_2(\kappa) \xi
\end{align}
which recovers \eqref{eq:phig} in Result \ref{conj:main:gaussian_equivalence}.

%% file: sections/appendix/replica.tex
\section{Derivation of Result \ref{res:asymptotics}}
\label{sec:app:replica}
In this Appendix, we detail the derivation of Result \ref{res:asymptotics}. The derivation leverages the replica method from statistical physics \cite{parisi1979toward, parisi1983order} in its replica-symmetric formulation, in conjunction with Result \ref{conj:main:gaussian_equivalence}. On a technical level, the derivation builds upon the formal equivalence between the distribution of the equivalent features $\vec \varphi^g(x)$ \eqref{eq:phig} with a mixture of Gaussians with infinitely many $\kappa-$indexed clusters, similarly to the superstatistical approach of \cite{adomaityte2023high, adomaityte2023classification} in another setting.

% Result \ref{res:asymptotics} thus rephrases the high-dimensional learning problem \eqref{eq:risk} in terms of a finite set of scalar summary statistics. 
% While we state Result \ref{res:asymptotics} for square loss and $\ell_2$ regularization for clarity, we conjecture a sharp characterization for generic convex loss $\ell$ in Appendix \ref{sec:app:replica}. 

%%%%%%%%%%%%%%%%%%%%%%%%%%%%%%%%%%%%%%%%%%%%%
First observe that for any test function $\phi(\hat{\vec a})$ of the minimizer $\hat{\vec a}$ of \eqref{eq:ERM_a}, 
\begin{align}
\label{eq:test_func}
    \phi(\hat{\vec a})=\underset{\omega \to \infty}{ \lim } \mathbb{E}_{\mathcal{D}} \frac{1}{Z}\int d \vec a \phi(\vec a)e^{-\omega \mathcal{R}_{\lambda}[\vec a]},
\end{align}
where we remind that $R[\vec a]$ denotes the empirical risk \eqref{eq:ERM_a}, and we denoted
\begin{align}
    Z\equiv \int d \vec a e^{-\omega \mathcal{R}_{\lambda}[\vec a]}
\end{align}
the normalization factor, also known as the \textit{partition function} in statistical physics. It is therefore important to the generating function associated with the measure \eqref{eq:test_func}, namely $\mathbb{E}\ln Z$. Such computations can be addressed using the \textit{replica} method from statistical physics \cite{parisi1979toward, parisi1983order}, leveraging the identity
\begin{align}
    \ln Z=\underset{s\to 0}{\lim}\frac{Z^s-1}{s}.
\end{align}
The backbone of the derivation thus lies in the computation of $\mathbb{E} Z^s$.

\subsection{Reminder of setting and notations}
Because of Result \ref{conj:main:gaussian_equivalence}, we take as a starting point that one can assume the sRF features $\vec z=\vec \varphi(x)$ \eqref{eq:phi} are distributed as 
\begin{align}
    \label{eq:data}
    \vec z&=\mu_0(\kappa) \vec 1_p+\mu_1(\kappa) \W \vec x +\mu_2(\kappa)\vec \xi\notag\\
    &=\mu_0(\kappa) \vec 1_p+\sfrac{1}{\sqrt{d}}\mu_1(\kappa)\kappa \W \vec v +\mu_1(\kappa) \W \Pi^\perp \vec x+\mu_2(\kappa) \vec \xi
\end{align}
where $\kappa\equiv \frac{\vec v^\top \vec z}{\sqrt{d}}\sim\mathcal{N}(0,1), \vec u= \vec 1_d, \vec x\sim\mathcal{N}(0,\mathbb{I}_d), \vec \xi\sim\mathcal{N}(0,\mathbb{I}_p) $. We remind that

\begin{align}
\label{eq:app:mus}
\mu_0(\kappa) &=  \Ea_{w}{\sigma(w+ r \kappa)}    \\
\mu_1(\kappa)&=  \frac{1}{c}\Ea_{w}{w\sigma(w+r \kappa)}    \\
\mu_2(\kappa) &= \sqrt{\Ea_{w}{\sigma^2(w+ r \kappa)-c(\mu_1(\kappa))^2-(\mu_0(\kappa))^2}},
\end{align}
where expectations bear over $w\sim\mathcal{N}(0,c)$. Finally, we consider a generic target function
\begin{align}
    f_\star(\vec x)=\sigma_\star(\kappa, \sfrac{ \vec \theta^\top \Pi^\perp \vec x}{\sqrt{d}}).
\end{align}
Note that single-index models considered in the main text constitute a special case of this family of functions, which we consider here for generality.
We look at the empirical risk minimization over $\vec a$ of 
\begin{align}
    \label{eq:risk2}
    R[\vec a]=\sum\limits_{\mu=1}^n\ell(f_\star(\vec x^\mu),\sfrac{\vec a^\top \vec z^\mu}{\sqrt{p}})+g(\vec a),
\end{align}
for generic convex loss $\ell$ and $\ell_2$ regularizer $g(\cdot)=\sfrac{\lambda}{2}\lVert \cdot\lVert^2$. We shall provide a detailed derivation in this generic case and only specialize to the particular case of square loss addressed in Result \ref{res:asymptotics} at the end of the derivation.

\subsection{Replica computation}
The replicated partition function reads
\begin{align}
    Z^s=\int \prod\limits_{r=1}^s d\vec a_r e^{-\omega g(\vec a_r)}\left[\underbrace{
    \mathbb{E}_{x} \prod\limits_{r=1}^s e^{-\omega \ell(f_\star(\vec x),\sfrac{\vec a_r^\top \vec z}{\sqrt{p}})}}_{(\star)}
    \right]^n
\end{align}
We now expand $(\star)$.
\begin{align}
    (\star)&=\mathbb{E}_{\kappa, \vec x^\perp, \vec \xi} \prod\limits_{r=1}^s e^{-\omega \ell(\sigma_\star(\kappa, \lambda_\star),\mu_0(\kappa) m_r+\mu_1(\kappa)\kappa \zeta_r+\mu_1(\kappa)g_r+\mu_2(\kappa)h_r)}
\end{align}
We defined the summary statistics
\begin{align}
    m_r=\frac{\vec a_r^\top \vec 1_p}{\sqrt{p}},&&
    \zeta_r=\frac{\vec a_r^\top \W \vec v}{\sqrt{dp}}, &&
    q^1_{r r'}=\frac{\vec a_r^\top \Omega \vec a_{r'}}{p},&&q^2_{rr'}=\frac{\vec a_r^\top \vec a_{r'}}{p},
\end{align}
and local fields
\begin{align}
    g_r=\frac{\vec a_r^\top \W \Pi^\perp \vec x}{\sqrt{p}},&&
    h_r=\frac{\vec a_r^\top \vec \xi}{\sqrt{p}}, &&\lambda_\star =\frac{\vec \theta^\top \Pi^\perp \vec x}{\sqrt{d}}
\end{align}
where
\begin{align}
    \Omega=\W \Pi^\perp \W^\top.
\end{align}
One can then rewrite $(\star)$ more compactly as
\begin{align}
    (\star)=\mathbb{E}_{\kappa}\left[\mathbb{E}_{\lambda_\star,\{\lambda_r\}_{r=1}^s} e^{-\omega\sum\limits_{r=1}^s \ell(\sigma_\star(\kappa,\lambda_\star),\mu_0(\kappa) m_r+\mu_1(\kappa)\kappa \zeta_r+\lambda_r)}\right]
\end{align}
where the fields $\lambda_\star,\lambda_r$ are Gaussian with statistics
\begin{align}
    &\langle \lambda_r\lambda_{r'}\rangle=\mu_1(\kappa)^2q^1_{rr'}+\mu_2(\kappa)^2q^2_{rr'}\notag\\
    &\langle \lambda_\star^2\rangle=\frac{\vec \theta^\top\Pi^\perp \vec \theta}{d}\equiv \rho\notag\\
    &\langle \lambda_r\lambda_\star\rangle=\mu_1(\kappa)\frac{\vec a_r^\top \W\Pi^\perp \vec \theta}{\sqrt{pd}}\equiv \mu_1(\kappa)\psi_r
\end{align}
Introducing Dirac's deltas in their Fourier form to enforce the definitions of the summary statistics $m_r,\zeta_r,q^{1,2}_{rr'},\psi_r$, the replicated partition function then reads
\begin{align}
    Z^s=&\int \prod\limits_{r=1}^s dm_rd\hat{m}_rd\zeta_rd\hat{\zeta}_rd\psi_rd\hat{\psi}_r\prod\limits_{r,r'}dq^1_{rr'}d\hat{q}^1_{rr'}
    dq^2_{rr'}d\hat{q}^2_{rr'}\underbrace{ e^{-\sqrt{p}\sum\limits_r m_r\hat{m}_r
    -\sqrt{dp}\sum\limits_r (\zeta_r\hat{\zeta}_r+\psi_r\hat{\psi}_r)
    -p\sum\limits_{a\le b}(q^1_{rr'}\hat{q}^1_{rr'}+q^2_{rr'}\hat{q}^2_{rr'})}}_{e^{s \omega p \Psi_t}}
    \notag\\
    &\underbrace{\int \prod\limits_{r}d \vec a_r e^{-\omega \sum\limits_r g(\vec a_r)}e^{+\sum\limits_{r=1}^s \vec a_r^\top(\hat{m}_r  \vec 1_p+\hat{\zeta}_r \W \vec v+\hat{\psi}_r \W\Pi^\perp \vec \theta ) +\sum\limits_{r\le r'} (\hat{q}^1_{rr'} \vec w_r^\top \Omega \vec w_{r'}+\hat{q}^2_{rr'} \vec w_r^\top \vec w_b)}}_{e^{s \omega p \Psi_w}}\notag\\
    &\underbrace{\left[\mathbb{E}_{\kappa}\left(\mathbb{E}_{\lambda_\star,\{\lambda_r\}_{r=1}^s} e^{-\omega\sum\limits_{r=1}^s \ell(\sigma_\star(\kappa,\lambda_\star),\mu_0(\kappa) m^a+\mu_1(\kappa)\kappa \zeta^a+\lambda_r)}\right)\right]^n}_{e^{\sfrac{\alpha}{\beta} s\omega p\Psi_y}}
\end{align}
We will refer to these three terms respectively as the trace, entropic, and energetic terms. Note that all terms in the exponents are a priori $\Theta_d(d)$, thus large. It is therefore possible to compute the integral over the summary statistics $m_r,\zeta_r,q^{1,2}_{rr'},\psi_r$ and respective conjugate variables using Laplace's method, thereby reducing the problem to finding the extremum of $s\Psi_t+s\Psi_w+\alpha s \Psi_y$. 

\subsection{Replica symmetric ansatz}
The extremization remains challenging on several levels, notably because it involves $s^2+5s$ variables, and the replica method requires to take the limit $s\to 0$. To make the problem tractable, we will seek the extremum assuming it has the \textit{replica-symmetric} form 
\begin{align}
    &\forall 1\le r,r'\le s,~~q^\iota_{rr'}=\delta_{rr'}(r^\iota-q^\iota)+q^\iota\notag\\
    &\forall 1\le r\le s,~~m_r=m\notag\\
    &\forall 1\le r\le s,~~\zeta_r=\zeta\notag\\
    &\forall 1\le r\le s,~~\psi_r=\psi
\end{align}
for $\iota=1,2$, and similarly for the hat variables:
\begin{align}
    &\forall 1\le r,r'\le s,~~\hat{q}^\iota_{ab}=\delta_{ab}(-\frac{1}{2} \hat{r}^\iota -\hat{q}^\iota)+\hat{q}^\iota\notag\\
    &\forall 1\le r\le s,~~\hat{m}_r=\hat{m}\notag\\
    &\forall 1\le r\le s,~~\hat{\zeta}_r=\hat{\zeta}\notag\\
    &\forall 1\le r\le s,~~\hat{\psi}_r=\hat{\psi}
\end{align}
Like in e.g. \cite{loureiro2021learning}, we also introduce the variance order parameters
\begin{align}
    V_\iota=r_\iota-q_\iota,&& \hat{V}_\iota=\hat{r}_\iota+\hat{q}_\iota.
\end{align}

\subsection{Computing the free energy}
\paragraph{Entropic term}
The computation of the entropic term follows from the derivation of \cite{aubin2020generalization}, with but minor changes. The entropic term thus reads, 
\begin{align}
    p\omega\Psi_w=\mathbb{E}_{\vec \xi}\ln\int d\vec a e^{-\omega g(\vec a)-\frac{1}{2}\vec a^\top (\hat{V}_1\Omega+\hat{V}_2\mathbb{I}_p)\vec a+\vec a^\top \left(
    \hat{m}1_p+ \hat{\zeta}\W \vec v+ \hat{\psi}\W\Pi^\perp \vec \zeta+\left(\hat{q}_1\Omega+\hat{q}_2\mathbb{I}_p\right)^{\sfrac{1}{2}} \vec \xi
    \right)}
\end{align}
In the $\omega\to\infty$ limit, rescaling the conjugate statistics as
\begin{align}
\hat{q}_\iota\to\omega^2\hat{q}_\iota, &&\hat{m},\hat{V}_\iota,\hat{\zeta},\hat{\psi}\to\omega\hat{m},\omega\hat{V}_\iota,\omega\hat{\zeta},\omega\hat{\psi}, 
\end{align}
the entropic potential can finally be rewritten in the $\omega\to\infty$ limit as
\begin{align}
    \Psi_w=-\mathbb{E}_{\vec \xi}\mathcal{M}_g(\vec \xi)
\end{align}
where we define the Moreau envelope
\begin{align}
    \mathcal{M}_g(\vec \xi)=\underset{\vec a}{\inf}\left\{
    g(\vec a)+\frac{1}{2} \vec a^\top (\hat{V}_1\Omega+\hat{V}_2\mathbb{I}_p) \vec a-a^\top \left(
    \hat{m} \vec 1_p+ \hat{\zeta}\W \vec v+ \hat{\psi}\W\Pi^\perp \vec \theta+\left(\hat{q}_1\Omega+\hat{q}_2\mathbb{I}_p\right)^{\sfrac{1}{2}} \vec \xi
    \right)
    \right\}
\end{align}
For the case $g=\sfrac{\lambda}{2}\lVert\cdot\lVert^2$, this simplifies to 
\begin{align}
    \Psi_w=\frac{1}{2p}\Tr[\left(
    \hat{m}^2 \vec 1_p \vec 1_p^\top+\hat{\zeta}^2 \W \vec v \vec v^\top \W^\top +\hat{\psi}^2 \W\Pi^\perp \vec \theta \vec \theta^\top\Pi^\perp \W^\top +\hat{q}_1\Omega+\hat{q}_2\mathbb{I}_d
    \right)\left(
    \hat{V}_1\Omega+\hat{V}_2\mathbb{I}_d+\lambda \mathbb{I}_d
    \right)^{-1}].
\end{align}
We neglected cross-terms of the form $\vec{1}_p\vec{v}^\top W^\top,\vec{1}_p\vec{\theta}^\top \Pi^\top W^\top, W\vec{v} \vec{\theta}^\top \Pi^\top W^\top $ in the numerator, which are expected to be asymptotically subleading. For example, one expects
\begin{align}
    \frac{1}{p}\Tr[\vec{1}_p\vec{v}^\top W^\top \left(
    \hat{V}_1\Omega+\hat{V}_2\mathbb{I}_d+\lambda \mathbb{I}_d
    \right)^{-1}]&=\frac{1}{p}\vec{v}^\top W^\top \left(
    S^{-1}+\frac{\hat{V}_1}{d}\frac{S^{-1}\W vv^\top \W^\top S^{-1}}{1-\frac{\hat{V}_1}{d}\Tr[\W vv^\top \W^\top S^{-1}]}
    \right) \vec{1}_p\notag\\
    &=O_d(\sfrac{1}{\sqrt{d}}),
\end{align}
where we introduced the shorthand $S=\hat{V}_1WW^\top +\hat{V}_2\mathbb{I}_p+\lambda\mathbb{I}_p$ and used the Sherman-Morrison lemma. The last scaling follows from the fact that since the vectors $W\vec{v}, \vec{1}_p$ are weakly correlated, their scalar product --with the uncorrelated matrix $S^{-1}$ in the middle-- should be $\vec{v}^\top W^\top S^{-1}\vec{1}_p=O_d(\sfrac{1}{\sqrt{d}})$. A similar reasoning holds for the other cross-terms.

\paragraph{Energetic term} The energetic term is the same as for a Gaussian mixture computation as in e.g. \cite{mignacco2020role, loureiro2021learning2, pesce2023gaussian,cui2023high} for Gaussian mixture distributions, with the $\mathbb{E}_\kappa$ playing the role of the sum over clusters. In other words, the equivalent feature map \eqref{eq:phig} is formally equivalent to an \textit{infinite mixture} of Gaussian clusters indexed by the projection $\kappa$ on the spike $\vec v$. Note that the formal equivalence to infinite Gaussian mixtures was already studied in \cite{adomaityte2023high, adomaityte2023classification}, in the context of learning from heavy-tailed data.
Exploiting this formal equivalence, the expectation over $\{\lambda_a\}$ can be carried out and give a Moreau envelope in the $\omega \to \infty$ limit, yielding
\begin{align}
\label{eq:Psiy}
    \Psi_y=-\mathbb{E}_{\kappa,\xi,y}\mathcal{M}(\xi,\kappa,y)
\end{align}
where $\xi\sim\mathcal{N}(0,1)$ and 
\begin{align}
    \mathcal{M}(\xi,\kappa,y)=\underset{x}{\inf}\left\{\frac{1}{2V(\kappa)}(x-\mu_1(\kappa)\psi y-\sqrt{\rho q(\kappa) -\mu_1(\kappa)^2\psi^2}\xi)^2+\ell(\sigma_\star(\kappa,y),\mu_0(\kappa)m+\mu_1(\kappa)\kappa\zeta+x)\right\}
\end{align}
we noted
\begin{align}
    q(\kappa)=\mu_1(\kappa)^2q_1+\mu_2(\kappa)^2q_2,
    && V(\kappa)=\mu_1(\kappa)^2V_1+\mu_2(\kappa)^2V_2.
\end{align}
In \eqref{eq:Psiy}, all expectations bear over zero-mean, unit variance Gaussian variables. Note that $q(\kappa), V(\kappa)$ are the exact equivalents of $q_k, V_k$ in e.g. \cite{loureiro2021learning2} or \cite{cui2023high}, i.e. the overlaps associated to cluster index $\kappa$. As previously mentioned, the equivalent map \eqref{eq:phig} formally coincides with an infinite Gaussian mixture, with $\kappa$ indexing the cluster.

\paragraph{Trace term} Note that in contrast to e.g. \cite{mignacco2020role,loureiro2021learning2, cui2023high}, in the trace term $\Psi_t$, the $\hat{m}m$ term is multiplied by $\sqrt{d}$ instead of $d$. This is due to the fact that the cluster centroids, whose role is here played by $u=1_p$, is of norm $\sqrt{d}$, whereas \cite{mignacco2020role}\cite{loureiro2021learning2} consider Gaussian clusters separated by centroids of norm $\Theta_d(1)$. The computation of the trace term however follows identical steps as e.g. \cite{loureiro2021learning}, yielding
\begin{align}
    \Psi_t&=-\frac{1}{2}(\hat{V}_1q_1-\hat{q}_1V_1)-\frac{1}{2}(\hat{V}_2q_2-\hat{q}_2V_2)+\frac{1}{\sqrt{d}}m\hat{m}+\frac{1}{\sqrt{\beta}}\psi\hat{\psi}+\frac{1}{\sqrt{\beta}}\zeta\hat{\zeta}\notag\\
    &=-\frac{1}{2}(\hat{V}_1q_1-\hat{q}_1V_1)-\frac{1}{2}(\hat{V}_2q_2-\hat{q}_2V_2)+\frac{1}{\sqrt{\beta}}\psi\hat{\psi}+\frac{1}{\sqrt{\beta}}\zeta\hat{\zeta}
\end{align}
Comparing to \cite{loureiro2021learning2}, the $\hat{m}m$ term in the free energy is thus vanishing asymptotically.

\paragraph{Free energy}
Putting everything together, the free energy is 
\begin{align}
    f=&\underset{q_\iota,V_\iota,m,\psi,\zeta, \hat{q}_\iota,\hat{V}_\iota,\hat{m},\hat{\psi},\hat{\zeta}}{\mathrm{extr}}\Bigg\{-\frac{1}{2}(\hat{V}_1q_1-\hat{q}_1V_1)-\frac{1}{2}(\hat{V}_2q_2-\hat{q}_2V_2)+\frac{1}{\sqrt{\beta}}\psi\hat{\psi}+\frac{1}{\sqrt{\beta}}\zeta\hat{\zeta}\notag\\
    &-\frac{1}{2p}\Tr[\left(
    \hat{m}^21_p1_p^\top+\hat{\zeta}^2 \W vv^\top \W^\top +\hat{\psi}^2 \W\Pi^\perp\vec \theta\vec \theta^\top\Pi^\perp \W^\top +\hat{q}_1\Omega+\hat{q}_2\mathbb{I}_d
    \right)\left(
    \hat{V}_1\Omega+\hat{V}_2\mathbb{I}_d+\lambda \mathbb{I}_d
    \right)^{-1}]\notag\\
    &+\frac{\alpha}{\beta} \mathbb{E}_{\kappa,\xi}\mathcal{M}(\xi,\kappa)\Bigg\}
\end{align}

The free energy can be simplified, noting that the zero-gradient (saddle-point) condition on $\hat{m}$ can be written as
\begin{align}
    0\overset{!}{=} \hat{m}\Tr[1_p1_p^\top \left(
    \hat{V}_1\Omega+\hat{V}_2\mathbb{I}_d+\lambda \mathbb{I}_d
    \right)^{-1}],
\end{align}
which imposes $\hat{m}=0$. The free energy now simplifies to 
\begin{align}
    f=&\underset{q_\iota,V_\iota,m,\psi,\zeta, \hat{q}_\iota,\hat{V}_\iota,\hat{\psi},\hat{\zeta}}{\mathrm{extr}}\Bigg\{-\frac{1}{2}(\hat{V}_1q_1-\hat{q}_1V_1)-\frac{1}{2}(\hat{V}_2q_2-\hat{q}_2V_2)+\frac{1}{\sqrt{\beta}}\psi\hat{\psi}+\frac{1}{\sqrt{\beta}}\zeta\hat{\zeta}\notag\\
    &-\frac{1}{2p}\Tr[\left(
    \hat{\zeta}^2 \W vv^\top \W^\top +\hat{\psi}^2 \W\Pi^\perp\vec \theta \vec \theta^\top\Pi^\perp \W^\top +\hat{q}_1\Omega+\hat{q}_2\mathbb{I}_d
    \right)\left(
    \hat{V}_1\Omega+\hat{V}_2\mathbb{I}_d+\lambda \mathbb{I}_d
    \right)^{-1}]\notag\\
    &+\frac{\alpha}{\beta} \mathbb{E}_{\kappa,\xi}\mathcal{M}(\xi,\kappa)\Bigg\}
\end{align}

Let us massage the expression in the trace further, by explicitly replacing $\Omega$ by its definition, namely $\Omega=\W\W^\top -\sfrac{1}{d}\W vv^\top\W^\top$. The term in the trace then becomes
\begin{align}
    (a)\!\!\equiv\!\!\frac{1}{p}\!\!\Tr[\!\!\left(
    (\hat{\zeta}^2\!\!-\!\!\sfrac{\hat{q}_1}{d}) \W vv^\top \W^\top \!\!+\hat{\psi}^2 \W\Pi^\perp\vec \theta \vec \theta^\top\Pi^\perp \W^\top \!\!+\!\!\hat{q}_1\W\W^\top\!\!+\!\!\hat{q}_2\mathbb{I}_d
    \right)\!\!\left(
    \hat{V}_1\W\W^\top\!\!-\!\!\sfrac{\hat{V}_1}{d}\W vv^\top \W+\hat{V}_2\mathbb{I}_d+\lambda \mathbb{I}_d
    \right)^{-1}]
\end{align}
The $\sfrac{\hat{q}_1}{d}$ in the numerator can be safely neglected as a small correction to $\hat{\zeta}^2$. On the other hand, one has to take care of the spike in the denominator using the Sherman-Morrison lemma. Using the shorthand
\begin{align}
    S\equiv \hat{V}_1\W\W^\top+\hat{V}_2\mathbb{I}_p+\lambda \mathbb{I}_p
\end{align}
one can write the denominator as
\begin{align}
    (S-\sfrac{\hat{V}_1}{d}\W vv^\top \W^\top)^{-1}=S^{-1}+\frac{\hat{V}_1}{d}\frac{S^{-1}\W vv^\top \W^\top S^{-1}}{1-\frac{\hat{V}_1}{d}\Tr[\W vv^\top \W^\top S^{-1}]}.
\end{align}
The trace term $(a)$ thus decomposes as 
\begin{align}
    (a)=&\frac{1}{p}\Tr[\left(
    \hat{\zeta}^2 \W vv^\top \W^\top +\hat{\psi}^2 \W\Pi^\perp\vec \theta \vec \theta^\top\Pi^\perp \W^\top +\hat{q}_1\W\W^\top+\hat{q}_2\mathbb{I}_p
    \right)S^{-1}]\notag\\
    &+\underbrace{\frac{\hat{V}_1}{pd}\frac{\Tr[\left(
    \hat{\zeta}^2 \W vv^\top \W^\top +\hat{\psi}^2 \W\Pi^\perp\vec \theta \vec \theta^\top\Pi^\perp \W^\top +\hat{q}_1\W\W^\top+\hat{q}_2\mathbb{I}_p
    \right)S^{-1}\W vv^\top \W^\top S^{-1}]}{1-\frac{\hat{V}_1}{d}\Tr[\W vv^\top \W^\top S^{-1}]}}_{(b)}
\end{align}
We turn to the simplification of the numerator of $(b)$, sequentially examining each of the three terms composing it. First, 
\begin{align}
    \hat{\psi}^2 \W\Pi^\perp\vec \theta \vec \theta^\top\Pi^\perp \W^\top S^{-1}\W vv^\top \W^\top S^{-1}&=\hat{\psi}^2\left(
    (\Pi^\perp\theta)^\top  W^\top S^{-1}W v
    \right)^2=O_d(d),
\end{align}
where we used that $\Pi^\perp \theta, v$ are orthogonal, so the squared scalar product of these vectors, with the uncorrelated matrix in between, should be asymptotically at most $O_d(\sqrt{d})$. Turning to the second term:
\begin{align}
    \hat{q}_1\W\W^\top S^{-1}\W vv^\top \W^\top S^{-1}&=\hat{q}_1\lVert \W^\top S^{-1}\W v\lVert^2=O_d(d).
\end{align}
Finally
\begin{align}
    \hat{\zeta}^2 \W vv^\top \W^\top  S^{-1}\W vv^\top \W^\top S^{-1}=\hat{\zeta}^2 \left(
    v^\top \W^\top  S^{-1}\W v
    \right)^2=\Theta_d(d^2).
\end{align}
In short, only the third term yields a $\Theta_d(1)$ contribution in $(b)$, i.e.
\begin{align}
    (b)=o_d(1)+\frac{\hat{\zeta}^2\hat{V}_1}{pd}\frac{\Tr[\W vv^\top \W^\top S^{-1}]^2}{1-\frac{\hat{V}_1}{d}\Tr[\W vv^\top \W^\top S^{-1}]}
\end{align}
Finally, the $(a)$ term can be written be written in fully asymptotic form, by introducing the limiting distribution $\nu(\varrho,\tau\pi)$ of Assumption \ref{ass:assumption}:
\begin{align}
    (a)=&\int d\nu(\varrho,\tau,\pi)\frac{
    \hat{\zeta}^2 \tau^2\varrho +\hat{\psi}^2 \varrho \pi^2 +\hat{q}_1\varrho+\hat{q}_2
    }{
    \hat{V}_1\varrho+\hat{V}_2+\lambda 
    }+\beta\hat{\zeta}^2\hat{V}_1\frac{\left(\int d\nu(\varrho,\tau,\pi)
    \frac{\tau^2\varrho }{\hat{V}_1\varrho+\hat{V}_2+\lambda }
    \right)^2}{1-\beta\hat{V}_1\int d\nu(\varrho,\tau,\pi)
    \frac{\tau^2\varrho }{\hat{V}_1\varrho+\hat{V}_2+\lambda }},
\end{align}

thereby yielding the expression for the free energy
\begin{align}
\label{eq:free_energy}
    f=&\underset{q_\iota,V_\iota,m,\psi,\zeta, \hat{q}_\iota,\hat{V}_\iota,\hat{\psi},\hat{\zeta}}{\mathrm{extr}}\Bigg\{-\frac{1}{2}(\hat{V}_1q_1-\hat{q}_1V_1)-\frac{1}{2}(\hat{V}_2q_2-\hat{q}_2V_2)+\frac{1}{\sqrt{\beta}}\psi\hat{\psi}+\frac{1}{\sqrt{\beta}}\zeta\hat{\zeta}+\frac{\alpha}{\beta} \mathbb{E}_{\kappa,\xi}\mathcal{M}(\xi,\kappa)\notag\\
    &-\frac{1}{2}\int d\nu(\varrho,\tau,\pi)\frac{
    \hat{\zeta}^2 \tau^2\varrho +\hat{\psi}^2 \varrho \pi^2 +\hat{q}_1\varrho+\hat{q}_2
    }{
    \hat{V}_1\varrho+\hat{V}_2+\lambda 
    }-\beta\frac{\hat{\zeta}^2\hat{V}_1}{2}\frac{\left(\int d\nu(\varrho,\tau,\pi)
    \frac{\tau^2\varrho }{\hat{V}_1\varrho+\hat{V}_2+\lambda }
    \right)^2}{1-\beta\hat{V}_1\int d\nu(\varrho,\tau,\pi)
    \frac{\tau^2\varrho }{\hat{V}_1\varrho+\hat{V}_2+\lambda }}\Bigg\}
\end{align}

\paragraph{Comments on the limiting density $\nu$}  Let us briefly comment on the form of the limiting distribution $\nu$. To build intuition, let us further relax in this paragraph only the strict normalization on the norm of the rows of $W$, and assume the rows are independently sampled from $\mathcal{N}(0,c\mathbb{I}_d)$. Finally, by the same token, let us assume $\vec v\sim\mathcal{N}(0,\mathbb{I}_d)$ and $\vec \theta \overset{d}{=}\gamma \vec v + \mathcal{N}(0,(1-\gamma^2)\mathbb{I}_d)$, so that they are Gaussian vectors with standard deviation $\sqrt{d}$ and average overlap $\gamma$. Note that relaxing the spherical constraint on all vectors, and allowing them to be sampled from Gaussian distributions with matching statistics, is expected to lead to an asymptotically equivalent problem, and is only considered in this paragraph to ease the discussion. In this setting, it is straightforward to see that the measure $\nu$ decomposes as the simple product measure
\begin{align}
    \nu= c\otimes \mu^\beta_{\rm MP} \times \check{\nu}
\end{align}
where $\mu^\beta_{\rm MP}$ denotes the Marcenko-Pastur density with aspect ratio $\beta$, and the notation $a\otimes \mu$ denotes the pushforward $(a\otimes \mu)(\lambda)=\mu(a \lambda)$. We denoted $\check{\nu}$ the joint distribution of the random variables $x, y-\gamma x$
for $x,y\sim\mathcal{N}(0,1)$ independently.

\subsection{Saddle-point equations}
We explicit the saddle-point equations associated to the optimization problem \eqref{eq:free_energy}. Define $x^\star$ as the  minimizer of the Moreau envelope, and the proximal
\begin{align}
    \prox(y,\kappa,\xi)=\frac{1}{V(\kappa)}(x^\star-\mu_1(\kappa)\psi y-\sqrt{ \rho q(\kappa)-\psi^2\mu_1(\kappa)^2}\xi).
\end{align}
The saddle-point equations read
\begin{align}
\label{eq:SP}
    &\begin{cases}
        q_1=\int d\nu(\varrho,\tau,\pi) \varrho\frac{\left(
\hat{q}_1 \varrho +\hat{q}_2
+\hat{\zeta}^2 \varrho\tau^2 +\hat{\psi}^2 \varrho\pi^2
\right)}{
\left(\lambda+\hat{V}_1\varrho +\hat{V}_2\right)^2
}\\
\qquad
-\beta\hat{\zeta}^2\frac{\left(\int d\nu(\varrho,\tau,\pi)
    \frac{\tau^2\varrho }{\hat{V}_1\varrho+\hat{V}_2+\lambda }\right)^2+\frac{1}{\beta}\int d\nu(\varrho,\tau,\pi) \frac{\tau^2\varrho^2}{(\lambda+\hat{V}_1\varrho +\hat{V}_2)^2}\left[
    \left(1-\beta\hat{V}_1\int d\nu(\varrho,\tau,\pi)
    \frac{\tau^2\varrho }{\hat{V}_1\varrho+\hat{V}_2+\lambda }\right)^2-1
    \right]}{\left(1-\beta\hat{V}_1\int d\nu(\varrho,\tau,\pi)
    \frac{\tau^2\varrho }{\hat{V}_1\varrho+\hat{V}_2+\lambda }\right)^2}
\\
    q_2=\int d\nu(\varrho,\tau,\pi) \frac{\left(
\hat{q}_1 \varrho +\hat{q}_2
+\hat{\zeta}^2 \varrho\tau^2 +\hat{\psi}^2 \varrho\pi^2
\right)}{
\left(\lambda+\hat{V}_1\varrho +\hat{V}_2\right)^2}
-\hat{\zeta}^2\int d\nu(\varrho,\tau,\pi) \frac{\tau^2\varrho}{(\lambda+\hat{V}_1\varrho +\hat{V}_2)^2}\left[
1-\frac{1}{\left(1-\beta\hat{V}_1\int d\nu(\varrho,\tau,\pi)
    \frac{\tau^2\varrho }{\hat{V}_1\varrho+\hat{V}_2+\lambda }\right)^2}
\right]
\\
V_1=\int d\nu(\varrho,\tau,\pi) \varrho\frac{1}{
\lambda+\hat{V}_1\varrho +\hat{V}_2}\\
V_2=\int d\nu(\varrho,\tau,\pi) \frac{1}{
\lambda+\hat{V}_1\varrho +\hat{V}_2}\\
m=\underset{m}{\inf} \mathbb{E}_{\kappa,y,\xi}\mathcal{M}\\
\zeta=\hat{\zeta}\sqrt{\beta} \int d\nu(\varrho,\tau,\pi) \varrho\tau^2\frac{1}{
\lambda+\hat{V}_1\varrho +\hat{V}_2}+\beta^{\sfrac{3}{2}}\hat{\zeta}\hat{V}_1\frac{\left(\int d\nu(\varrho,\tau,\pi)
    \frac{\tau^2\varrho }{\hat{V}_1\varrho+\hat{V}_2+\lambda }
    \right)^2}{1-\beta\hat{V}_1\int d\nu(\varrho,\tau,\pi)
    \frac{\tau^2\varrho }{\hat{V}_1\varrho+\hat{V}_2+\lambda }} \\
\psi=\hat{\psi}\sqrt{\beta} \int d\nu(\varrho,\tau,\pi) \varrho\pi^2\frac{1}{
\lambda+\hat{V}_1\varrho +\hat{V}_2}
    \end{cases}
    \\
&\begin{cases}
\hat{V}_1=-\frac{\alpha}{\beta}\mathbb{E}_{\kappa,y,\xi}\frac{\rho \mu_1(\kappa)^2}{\sqrt{\rho q(\kappa)-\psi^2\mu_1(\kappa)^2}}\prox(y,\kappa,\xi)\\
\hat{q}_1=\frac{\alpha}{\beta}\mathbb{E}_{\kappa,y.\xi}\mu_1(\kappa)^2\prox(y,\kappa,\xi)^2\\
\hat{V}_2=-\frac{\alpha}{\beta}\mathbb{E}_{\kappa,y,\xi}\frac{\rho \mu_2(\kappa)^2}{\sqrt{\rho q(\kappa)-\psi^2\mu_1(\kappa)^2}}\prox(y,\kappa,\xi)\\
\hat{q}_2=\frac{\alpha}{\beta}\mathbb{E}_{\kappa,y.\xi}\mu_2(\kappa)^2\prox(y,\kappa,\xi)^2\\
\hat{\zeta}=-\frac{\alpha}{ \sqrt{\beta}}\mathbb{E}_{\kappa,y,\xi}\kappa\mu_1(\kappa)\ell^\prime(\sigma_\star( \kappa,y),\mu_0(\kappa)m+\mu_1(\kappa)\kappa\zeta+ x^\star)
\\
\hat{\psi}=\frac{\alpha}{ \sqrt{\beta}}\mathbb{E}_{\kappa,y,\xi}\left(\mu_1(\kappa)y-\frac{\psi\mu_1(\kappa)^2}{\rho q(\kappa)-\psi^2\mu_1(\kappa)^2}\xi\right)\prox(y,\kappa,\xi)
\end{cases}
\end{align}

\subsection{Saddle point equations for the square loss}
These expressions simplify for the square loss. Define the minimizer $x^\star$ of the Moreau envelope for $\ell(y,z)=\sfrac{1}{2}(y-z)^2$:
\begin{align}
x^\star(\kappa,\xi)=\frac{V(\kappa)\left(\sigma_\star(\kappa,y)-\mu_0(\kappa)m-\mu_1(\kappa)\kappa \zeta-\mu_1(\kappa)\psi\right)+\sqrt{\rho q(\kappa)-\mu_1(\kappa)^2\psi^2}\xi+\mu_1(\kappa)\psi y }{1+V(\kappa)}.
\end{align}

The saddle point for $m$ admits a compact closed form expression. Writing the saddle point equation for $m$ indeed imposes that
\begin{align}
    0&=\partial_m\mathbb{E}_{\kappa,\xi}\mathcal{M}(\xi,\kappa)\notag\\
    &=\mathbb{E}_{\kappa,\xi}\left[\mu_0(\kappa)\left(
    f_\star(\kappa)-\mu_0(\kappa)m-\frac{V(\kappa)f^\star(\kappa)-V(\kappa)\mu_0(\kappa)m+\sqrt{q(\kappa)}\xi}{1+V(\kappa)}
    \right)\right],
\end{align}
from which it follows that
\begin{align}
    m \mathbb{E}_\kappa\left[
    \frac{\mu_0(\kappa)^2}{1+V(\kappa)}
    \right]=\mathbb{E}_\kappa\left[\frac{\mu_0(\kappa)f_\star(\kappa)}{1+V(\kappa)}\right],
\end{align}
i.e.
\begin{align}
    m=\frac{\mathbb{E}_\kappa\left[\frac{\mu_0(\kappa)f_\star(\kappa)}{1+V(\kappa)}\right]}{\mathbb{E}_\kappa\left[
    \frac{\mu_0(\kappa)^2}{1+V(\kappa)}
    \right]}
\end{align}
The saddle-point equations thus read
\begin{align}
    &\begin{cases}
          q_1=\int d\nu(\varrho,\tau,\pi) \varrho\frac{\left(
\hat{q}_1 \varrho +\hat{q}_2
+\hat{\zeta}^2 \varrho\tau^2 +\hat{\psi}^2 \varrho\pi^2
\right)}{
\left(\lambda+\hat{V}_1\varrho +\hat{V}_2\right)^2
}
\\
\qquad-\beta\hat{\zeta}^2\frac{\left(\int d\nu(\varrho,\tau,\pi)
    \frac{\tau^2\varrho }{\hat{V}_1\varrho+\hat{V}_2+\lambda }\right)^2+\frac{1}{\beta}\int d\nu(\varrho,\tau,\pi) \frac{\tau^2\varrho^2}{(\lambda+\hat{V}_1\varrho +\hat{V}_2)^2}\left[
    \left(1-\beta\hat{V}_1\int d\nu(\varrho,\tau,\pi)
    \frac{\tau^2\varrho }{\hat{V}_1\varrho+\hat{V}_2+\lambda }\right)^2-1
    \right]}{\left(1-\beta\hat{V}_1\int d\nu(\varrho,\tau,\pi)
    \frac{\tau^2\varrho }{\hat{V}_1\varrho+\hat{V}_2+\lambda }\right)^2}
\\
    q_2=\int d\nu(\varrho,\tau,\pi) \frac{\left(
\hat{q}_1 \varrho +\hat{q}_2
+\hat{\zeta}^2 \varrho\tau^2 +\hat{\psi}^2 \varrho\pi^2
\right)}{
\left(\lambda+\hat{V}_1\varrho +\hat{V}_2\right)^2}
-\hat{\zeta}^2\int d\nu(\varrho,\tau,\pi) \frac{\tau^2\varrho}{(\lambda+\hat{V}_1\varrho +\hat{V}_2)^2}\left[
1-\frac{1}{\left(1-\beta\hat{V}_1\int d\nu(\varrho,\tau,\pi)
    \frac{\tau^2\varrho }{\hat{V}_1\varrho+\hat{V}_2+\lambda }\right)^2}
\right]\\
V_1=\int d\nu(\varrho,\tau,\pi) \varrho\frac{1}{
\lambda+\hat{V}_1\varrho +\hat{V}_2}\\
V_2=\int d\nu(\varrho,\tau,\pi) \frac{1}{
\lambda+\hat{V}_1\varrho +\hat{V}_2}\\
m=\frac{1}{\mathbb{E}_\kappa\left[\frac{\mu_0(\kappa)^2}{1+V(\kappa)}\right]}
\mathbb{E}_{\kappa,y}\left[\frac{\mu_0(\kappa)(\sigma_\star(\kappa, y)-\mu_1(\kappa)\kappa\zeta)}{1+V(\kappa)}\right]\\
\zeta=\hat{\zeta}\sqrt{\beta} \int d\nu(\varrho,\tau,\pi) \varrho\tau^2\frac{1}{
\lambda+\hat{V}_1\varrho +\hat{V}_2}+\beta^{\sfrac{3}{2}}\hat{\zeta}\hat{V}_1\frac{\left(\int d\nu(\varrho,\tau,\pi)
    \frac{\tau^2\varrho }{\hat{V}_1\varrho+\hat{V}_2+\lambda }
    \right)^2}{1-\beta\hat{V}_1\int d\nu(\varrho,\tau,\pi)
    \frac{\tau^2\varrho }{\hat{V}_1\varrho+\hat{V}_2+\lambda }} \\
\psi=\hat{\psi}\sqrt{\beta} \int d\nu(\varrho,\tau,\pi) \varrho\pi^2\frac{1}{
\lambda+\hat{V}_1\varrho +\hat{V}_2}
    \end{cases}
    \\
&\begin{cases}
\hat{V}_1=\frac{\alpha}{\beta} \mathbb{E}_{\kappa}\frac{\rho\mu_1(\kappa)^2}{1+V(\kappa)}\\
\hat{q}_1=\frac{\alpha}{\beta}\mathbb{E}_{\kappa,y}\mu_1(\kappa)^2\frac{b(\kappa,y)^2+ q(\kappa)-\mu_1(\kappa)^2\psi^2}{\left(1+V(\kappa)\right)^2}\\
\hat{V}_2=\frac{\alpha}{\beta}\mathbb{E}_{\kappa}\frac{\rho\mu_2(\kappa)^2}{1+V(\kappa)}\\
\hat{q}_2=\frac{\alpha}{\beta} \mathbb{E}_{\kappa,y}\mu_2(\kappa)^2\frac{b(\kappa,y)^2+ q(\kappa)-\mu_1(\kappa)^2\psi^2}{\left(1+V(\kappa)\right)^2}\\
\hat{\zeta}=\frac{\alpha}{\sqrt{\beta}}\mathbb{E}_{\kappa,y}\kappa\mu_1(\kappa)\frac{b(\kappa,y)}{1+V(\kappa)}\\
\hat{\psi}=\frac{\alpha}{\sqrt{\beta}}\sqrt{\beta}\mathbb{E}_{\kappa,y}\frac{y\mu_1(\kappa)b(\kappa,y)+\psi\mu_1(\kappa)^2}{1+V(\kappa)}
\end{cases},
\end{align}
which recovers equations \eqref{eq:SP} from Result \ref{res:asymptotics}.

\subsection{Test error}
We conclude this Appendix by deriving the sharp asymptotic characterization for the test error. Using once more Result \ref{conj:main:gaussian_equivalence}, the test error reads
\begin{align}
    \epsilon_g&= \mathbb{E}_x\left(
    \sigma_\star(\kappa, \sfrac{\vec \theta^\top\Pi^\perp x}{\sqrt{d}})-\sfrac{\hat{a}^\top \varphi^g(x)}{\sqrt{p}}\right)^2\notag\\
    &= \mathbb{E}_{x,\xi}\left(
    \sigma_\star(\kappa, \sfrac{\vec \theta^\top\Pi^\perp x}{\sqrt{d}})-\mu_0(\kappa)\frac{\hat{a}^\top 1_p}{\sqrt{p}}-\mu_1(\kappa)\kappa\frac{\hat{a}^\top \W v}{\sqrt{dp}}-\mu_1(\kappa) \frac{\hat{a}^\top \W\Pi^\perp x}{\sqrt{p}}-\mu_2(\kappa)\xi
    \right)^2\notag\\
    &=\mathbb{E}_{\kappa, y,z}\left(
    \sigma_\star(\kappa, y)-\mu_0(\kappa)m-\mu_1(\kappa)\zeta-z
    \right)^2
\end{align}
with 
\begin{align}
    y,z\sim\mathcal{N}\left[0,\begin{pmatrix}
         \rho&\mu_1(\kappa)\psi\\
        \mu_1(\kappa)\psi& q(\kappa)    \end{pmatrix}\right].
\end{align}
The expression thus simplifies to 
\begin{align}
    \epsilon_g&=\mathbb{E}_{\kappa, y}\left[\left(
    \sigma_\star(\kappa, y)-\mu_0(\kappa)m-\mu_1(\kappa)\kappa\zeta-\frac{\mu_1(\kappa)\psi}{\sqrt{\rho}} y
    \right)^2 +q(\kappa)-\frac{\mu_1(\kappa)^2\psi^2}{\rho} \right]
\end{align}
where all expectations now bear on unit variance zero mean Gaussian variables.

\subsection{Additional discussion}
Before closing the present Appendix, we provide in this subsection further discussion of the results and highlight possible extensions.

\paragraph{Bias-- }Let us briefly sketch in this subsection how bias terms may be incorporated into the analysis. Consider the two-layer neural network
\begin{align}
    f_{W,\vec a,b}(\vec x)=\frac{1}{\sqrt{p}}\sum\limits_{i=1}^{p} a_{i} \sigma(\vec{w}_{i}^{\top}\vec x+b),
\end{align}
where $b\in\mathbb{R}$ is a bias that acts uniformly on all hidden-layer neurons. Then, conditional on $b$, the equivalent feature map still is of the form \eqref{eq:phig}, with coefficients
\begin{equation}
\begin{split}
\mu_0(\kappa,b) &=  \Ea_{z}{\sigma(z+ r \kappa+b)}    \\
\mu_1(\kappa,b)&=  \frac{1}{c}\Ea_{z}{z\sigma(z+r \kappa+b)}    \\
\mu_2(\kappa,b) &= \sqrt{\Ea_{z}{\sigma^2(z+ r \kappa+b)-c(\mu_1(\kappa))^2-(\mu_0(\kappa))^2}}.
\end{split}
\end{equation}
The analysis of the present Appendix still goes through, by treating $b$ as an additional network parameter to be replicated over. In fact, the only difference in introducing $b$ appears at the level of the Moreau envelope $\mathcal{M}$, in which all the $\mu_{0,1,2}(\kappa)$ coefficients are replaced by $\mu_{0,1,2}(\kappa+b)$. Finally, the corresponding saddle-point equation on $b$ should impose $\partial_b\mathcal{M}=0$, thereby imposing the value of the learned bias $\hat{b}$. This allows to improve the lower-bound \eqref{eq:lower_bound} as
\begin{align}
    \epsilon_g\ge  \underset{b_1,b_2,b}{\inf}\mathbb{E}_{\kappa}\Bigg[\Bigg(
    \sigma_\star(\kappa )-\mu_0(\kappa,b)b_1-\mu_1(\kappa,b)\kappa b_2
    \Bigg)^2\Bigg].
\end{align}
In other words, the bias allows an extra degree of freedom in the functional basis expressible by the network, thus allowing additional expressivity. We have addressed the case of uniform bias. While the case of generic non-uniform bias is unfortunately not presently in the reach of the techniques of this manuscript, the case where the components are untrained, but can take finitely many values, can be analyzed along similar lines as the case of non-uniform $u$. We briefly discuss this in Appendix \ref{app:non_uniform}.

\paragraph{Larger rank spikes-- } Let us now discuss the case where instead of a spike $\sfrac{1_pv^\top}{\sqrt{d}}$, we have a larger rank term $\sfrac{1_{p\times k} v^\top}{\sqrt{d}}$, with $k=\Theta_d(1)$ and where $v\in\mathbb{R}^{k\times d}$ is now a low-rank matrix instead of a vector. The equivalence of result \ref{conj:main:gaussian_equivalence} can be readily adapted as follows. Naming $v_1,..., v_k\in\mathbb{R}^d$ the $k$ columns of $v$, and conditioning on $\kappa_1,..., \kappa_k$ the $k$ projections of the data along these directions, the equivalent feature map \eqref{eq:phig} retains the same expression, with the coefficient $\mu^\prime_{0,1,2}(\kappa_1,..., \kappa_k)=\mu_{0,1,2}(\kappa_1+...+\kappa_k)$ now becoming multivariate.

\paragraph{Noisy target--} Finally, let us mention how one may adapt the analysis to a noisy target function $f_\star(x)+\mathcal{N}(0,1)$. The only point where this intervenes is during the evaluation of the energetic potential $\Psi_y$, which is identical to e.g. \cite{loureiro_learning_2021}, Appendix A. The only effect is to change the distribution of the variable $y$ in the expectation over the Moreau envelope, and we refer the interested reader to e.g. \cite{loureiro_learning_2021} for a full expression.

\begin{figure}
    \centering
    \includegraphics[scale=0.5]{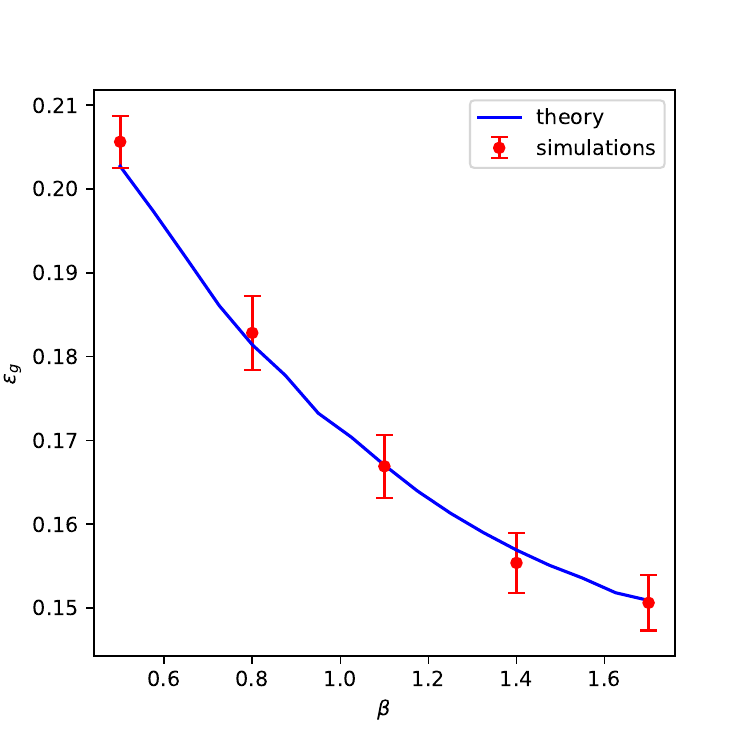}
    \caption{$\sigma=\tanh,\sigma_\star=\sin, \alpha_0=1.5,\alpha=1.2, \lambda=0.1$ Test error as a function of the network width $\beta$, as predicted by the theoretical characterization of Result \ref{res:asymptotics} (blue) or measure in numerical simulations in $d=5000$ (red); error bars represent one standard deviation over $20$ trials.}
    \label{fig:beta}
\end{figure}

\paragraph{Effect of network width--} What is the effect of the network width, as measured by $\beta$? On the one hand, large networks result in smaller spike strength $r$, as can be observed from Result \ref{res:GD_mapping}. On the other hand, it also allows for more overparametrization. The asymptotic characterization of Result \ref{res:asymptotics} allows to probe this question. Fig.\,\ref{fig:beta} shows that larger networks tend to achieve lower overall test error.

\color{black}

%% file: sections/appendix/non_uniform.tex
\section{Non-Uniform Readout Initialization}
\label{app:non_uniform}

\subsection{Setting}
In this Appendix, we briefly discuss the case where the readout layer is not initialized as a vector with equal components, $a^{(0)}=\sfrac{1_p}{\sqrt{p}}$, but the case where the components of $a^{(0)}$ can
 take values in a \textit{finite} vocabulary $V$, for instance $\{-1,0,+1\}$. We further assume the fraction of all letters are asymptotically  finite (i.e. the number of occurences of a letter $\sigma\in V$ in $a^{(0)}$, divided by $d$, tends to a well-defined finite value as $d\to\infty$). For simplicity, we provide a sketch of the derivation directly for the sRF model \eqref{eq:W}, and do not explicit for conciseness the mapping from the gradient-trained network thereto, which follows identical steps from the mapping detailed in Appendix \ref{app:GD}. More precisely, we consider directly a sRF with weights
 \begin{align}
     \label{eq:non_unif_W}
     \W+\frac{uv^\top }{\sqrt{d}}.
 \end{align}
 In contrast to the main text, we thus do not assume $u\propto 1_p$. We further also allow the norm of the rows of $\W$ to take different values $c(u_i)^{\sfrac{1}{2}}$ depending on the value taken by the component of $u$ with the corresponding index. This assumption originates from the form of the $\Delta$ matrix \eqref{eq:Delta}, which exhibits such row-wise norm variations matching that of $a^{(0)}$ (thus $u$ in the equivalent sRF). Finally, for simplicity and not to complicate the formulae, we assume perfect alignment with the target $\gamma=1$, and assume aspect ratio $p=d$, i.e. $\beta=1$. The generic case $0\le \gamma\le 1$ can be derived following identical steps to Appendix \ref{sec:app:replica}.
 
 \subsection{Sketch of derivation}
 
For non-uniform $u$, an equivalent feature map can be obtained along the same lines as for \ref{conj:main:gaussian_equivalence}, as detailed in Appendix \ref{app:CGET}. The sRF feature map has an asymptotically equivalent test error as the equivalent feature map
\begin{align}
    \varphi^g(x)= \sum\limits_{\sigma\in V}\mu_0^\sigma(\kappa)1_\sigma+\mu_1^\sigma(\kappa)\Pi_\sigma \W x+\mu_2^\sigma(\kappa)\Pi_\sigma \xi
\end{align}
where
\begin{align}
    1_\sigma=(\delta_{u_i,\sigma})_{i=1}^d&&\Pi_\sigma=\mathrm{diag}(1_\sigma)\in\mathbb{R}^{d\times d}
\end{align}
and ($z\sim\mathcal{N}(0,1)$ in the following)
\begin{align}
    &\mu_0^\tau(\kappa)=\mathbb{E}_{z\sim\mathcal{N}(0,c(\tau))}[\sigma(z+\tau\kappa)]\\
    &\mu_1^\tau(\kappa)=\frac{1}{c(\tau)}\mathbb{E}_{z\sim\mathcal{N}(0,c(\tau))}[z\sigma(z+\tau\kappa)]\\
    &\mu_0^\tau(\kappa)=\sqrt{\mathbb{E}_{z}[\sigma(z+\tau\kappa)^2]-\mu_0^\tau(\kappa)^2-c(\tau)\mu_1^\tau(\kappa)^2}
\end{align}
Note $\Pi_\sigma\Pi_\tau=\delta_{\sigma\tau}\Pi_\sigma$. 
Building on this equivalent map, the test error can by asymptotically characterized using the replica method, following the same lines as Appendix \ref{sec:app:replica}. Like in Appendix \ref{sec:app:replica}, we need to compute
\begin{align}
    (a)&=\mathbb{E}_{z, \xi} \prod\limits_{a=1}^s e^{-\omega \ell(\sigma_\star(\kappa),\sum\limits_\sigma\mu_0^\sigma(\kappa) m^a_\sigma+\mu_1^\sigma(\kappa)\kappa\zeta_\sigma^a+\mu_1^\sigma(\kappa)g^a_\sigma+\mu_2^\sigma(\kappa)h^a_\sigma)}
\end{align}
We defined the order parameters and local fields
\begin{align}
    m^a_\sigma=\frac{a_a^\top 1_\sigma}{\sqrt{d}}&&
    \zeta^a_\sigma=
    \frac{a_a^\top \Pi_\sigma \W v}{d}
    &&
    g^a_\sigma=\frac{a_a^\top \Pi_\sigma \W \Pi^\perp x}{\sqrt{d}}&&
    h^a_\sigma=\frac{a_a^\top \Pi_\sigma\xi}{\sqrt{d}}
\end{align}
Again, let us introduce the self-overlaps
\begin{align}
    (q^1_{\sigma\tau })_{ab}=\frac{a_a^\top \Omega_{\sigma\tau}a_b}{d}&&(q^2_{\sigma })_{ab}=\frac{a_a^\top \Pi_\sigma a_b}{d}
\end{align}
where we denoted 
\begin{align}
    \Omega_{\sigma\tau}=\Pi_\sigma \W\Pi \W^\top \Pi_\tau 
\end{align}
One can then rewrite $(a)$ more compactly as
\begin{align}
    (a)=\mathbb{E}_{\kappa}\left[\mathbb{E}_{\{\lambda_a\}_{a=1}^s} e^{-\omega\sum\limits_{a=1}^s \ell(\sigma_\star(\kappa),m^a(\kappa)+\lambda_a)}\right]
\end{align}
where the fields $\lambda_a$ are Gaussian with covariance
\begin{align}
    \langle \lambda_a\lambda_b\rangle=q_{ab}(\kappa)
\end{align}
where 
\begin{align}
    &q(\kappa)_{ab}=\sum\limits_{\sigma\tau}\mu_1^\sigma(\kappa)\mu_1^\tau(\kappa)q_{\sigma\tau,ab}^1+\sum\limits_{\sigma}\mu_2^\sigma(\kappa)^2 q^2_{\sigma,ab}\\
    &m^a(\kappa)=\sum\limits_{\sigma\in V}(\mu_0^\sigma(\kappa)m^a_\sigma(\kappa)+\mu_1^\sigma(\kappa)\kappa \zeta^a_\sigma(\kappa))
\end{align}

We again use the RS ansatz
form 
\begin{align}
    &\forall \sigma,\tau\in V, ~~\forall 1\le r,r'\le s,~~(q^\iota_{\sigma\tau})_{rr'}=\delta_{rr'}(r^\iota_{\sigma\tau}-q^\iota_{\sigma\tau})+q^\iota_{\sigma\tau}\notag\\
    &\forall \sigma \in V, ~~\forall 1\le r\le s,~~(m_\sigma)_r=m_\sigma\notag\\
    &\forall \sigma \in V, ~~\forall 1\le r\le s,~~(\zeta_\sigma)_r=\zeta_\sigma
\end{align}
for $\iota=1,2$, and similarly for the hat variables:
\begin{align}
    &\forall \sigma,\tau\in V, ~~\forall 1\le r,r'\le s,~~(\hat{q}^\iota_{\sigma\tau})_{rr^\prime}=\delta_{ab}(-\frac{1}{2} \hat{r}_{\sigma\tau}^\iota -\hat{q}_{\sigma\tau}^\iota)+\hat{q}_{\sigma\tau}^\iota\notag\\
    &\forall \sigma\in V, ~~\forall 1\le r\le s,~~(\hat{m}_\sigma)_r=\hat{m}_\sigma\notag\\
    &\forall \sigma\in V, ~~\forall 1\le r\le s,~~(\hat{\zeta}_\sigma)_r=\hat{\zeta}_\sigma
\end{align}

Finally, denoting
\begin{align}
    &q(\kappa)=\sum\limits_{\sigma\tau}\mu_1^\sigma(\kappa)\mu_1^\tau(\kappa)q_{\sigma\tau}^1+\sum\limits_{\sigma}\mu_2^\sigma(\kappa)^2 q^2_{\sigma}\\
    &V(\kappa)=\sum\limits_{\sigma\tau}\mu_1^\sigma(\kappa)\mu_1^\tau(\kappa)V_{\sigma\tau}^1+\sum\limits_{\sigma}\mu_2^\sigma(\kappa)^2 V^2_{\sigma}
\end{align}
one reaches along identical lines as Appendix \ref{sec:app:replica} the free energy
\begin{align}
    f=&-\frac{1}{2}\sum\limits_{\sigma\tau}(\hat{V}^1_{\sigma\tau}q^1_{\sigma\tau}-\hat{q}^1_{\sigma\tau}V^1_{\sigma\tau})-\frac{1}{2}\sum\limits_{\sigma}(\hat{V}^2_{\sigma}q^2_{\sigma}-\hat{q}^2_{\sigma}V^2_{\sigma})
    +\sum\limits_{\sigma} \zeta_\sigma\hat{\zeta}_\sigma
    \notag\\
    &-\frac{1}{2d}\Tr[\left(\sum\limits_{\sigma\le \tau}\hat{q}_{\sigma\tau}^1\Omega_{\sigma\tau}+\sum\limits_{\sigma\in V}\hat{q}^2_\sigma\Pi_\sigma +\sum\limits_\sigma \hat{\zeta}^2_\sigma \Pi_\sigma \W vv^\top \W^\top\Pi_\sigma 
    \right)\left(
    \sum\limits_{\sigma\le \tau}\hat{V}_{\sigma\tau}^1\Omega_{\sigma\tau}+\sum\limits_{\sigma\in V}\hat{V}^2_\sigma\Pi_\sigma+\lambda \mathbb{I}_d
    \right)^{-1}]\notag\\
    &+\alpha \mathbb{E}_{\kappa,\xi}\mathcal{M}(\xi,\kappa)
\end{align}
where
\begin{align}
    \Omega_{\sigma\tau}=\frac{1}{2}\Pi_\sigma \W\Pi \W^\top \Pi_\tau +\frac{1}{2}\Pi_\tau \W\Pi \W^\top \Pi_\sigma.
\end{align}
The expression of the Moreau is almost unchanged compared to Appendix  \ref{sec:app:replica}:
\begin{align}
    \mathcal{M}(\xi,\kappa)=\underset{x}{\inf}\frac{1}{2V(\kappa)}(x-\sqrt{q(\kappa)}\xi)^2+\ell(f^\star(\kappa),m(\kappa)+x),
\end{align}
where we noted
\begin{align}
    m(\kappa)=\sum\limits_{\sigma\in V}\mu_0^\sigma(\kappa)m_\sigma+\mu_1^\sigma(\kappa)\kappa\zeta_\sigma.
\end{align}

 The saddle point equations read, for the square loss $\ell(y,z)=\sfrac{1}{2}(y-z)^2$:

\begin{align}
    &\begin{cases}
        q^1_{\sigma\tau}=\frac{1}{d}\Tr[\Omega_{\sigma\tau}\left(\sum\limits_{\sigma\le \tau}\hat{V}_{\sigma\tau}^1\Omega_{\sigma\tau}+\sum\limits_{\sigma\in V}\hat{V}^2_\sigma\Pi_\sigma+\lambda \mathbb{I}_d
\right)^{-2}\left(\sum\limits_{\sigma\le \tau}\hat{q}_{\sigma\tau}^1\Omega_{\sigma\tau}+\sum\limits_{\sigma\in V}\hat{q}^2_\sigma\Pi_\sigma+\sum\limits_\sigma \hat{\zeta}^2_\sigma \Pi_\sigma \W vv^\top \W^\top\Pi_\sigma 
\right)
    ]\\
    q^2_\sigma=\frac{1}{d}\Tr[\Pi_\sigma\left(\sum\limits_{\sigma\le \tau}\hat{V}_{\sigma\tau}^1\Omega_{\sigma\tau}+\sum\limits_{\sigma\in V}\hat{V}^2_\sigma\Pi_\sigma+\lambda \mathbb{I}_d
\right)^{-2}\left(\sum\limits_{\sigma\le \tau}\hat{q}_{\sigma\tau}^1\Omega_{\sigma\tau}+\sum\limits_{\sigma\in V}\hat{q}^2_\sigma\Pi_\sigma+\sum\limits_\sigma \hat{\zeta}^2_\sigma \Pi_\sigma \W vv^\top \W^\top\Pi_\sigma 
\right)
    ]\\
V^1_{\sigma\tau}=\frac{1}{d}\Tr[\left(\sum\limits_{\sigma\le \tau}\hat{V}_{\sigma\tau}^1\Omega_{\sigma\tau}+\sum\limits_{\sigma\in V}\hat{V}^2_\sigma\Pi_\sigma+\lambda \mathbb{I}_d
\right)^{-1}\Omega_{\sigma\tau}]\\
V^2_\sigma=\frac{1}{d}\Tr[\left(\sum\limits_{\sigma\le \tau}\hat{V}_{\sigma\tau}^1\Omega_{\sigma\tau}+\sum\limits_{\sigma\in V}\hat{V}^2_\sigma\Pi_\sigma+\lambda \mathbb{I}_d
\right)^{-1}\Pi_\sigma]\\
\zeta_\sigma=\frac{\hat{\zeta}_\sigma}{d}\Tr[\left(\sum\limits_{\sigma\le \tau}\hat{V}_{\sigma\tau}^1\Omega_{\sigma\tau}+\sum\limits_{\sigma\in V}\hat{V}^2_\sigma\Pi_\sigma+\lambda \mathbb{I}_d
\right)^{-1}\Pi_\sigma \W vv^\top \W^\top\Pi_\sigma]
\\
m_\sigma=\frac{1}{\mathbb{E}_\kappa\left[\frac{\mu_0^\sigma(\kappa)^2}{1+V(\kappa)}\right]}
\mathbb{E}_\kappa\left[\frac{\mu_0^\sigma(\kappa)(f^\star(\kappa)-\sum\limits_{\tau\ne \sigma}\mu_0^\tau(\kappa)m_\tau
-\sum\limits_{\tau}\mu_1^\tau(\kappa)\kappa \zeta_\tau
)}{1+V(\kappa)}\right]
    \end{cases}
    \\
&\begin{cases}
\hat{V}^1_{\sigma\tau}=\alpha \mathbb{E}_{\kappa}\frac{\mu_1^\sigma(\kappa)\mu_1^\tau(\kappa)}{1+V(\kappa)}\\
\hat{q}^1_{\sigma\tau}=\alpha \mathbb{E}_{\kappa}\mu_1^\sigma(\kappa)\mu_1^\tau(\kappa)\frac{(f^\star(\kappa)-m(\kappa))^2+q(\kappa)}{\left(1+V(\kappa)\right)^2}\\
\hat{V}^2_{\sigma}=\alpha \mathbb{E}_{\kappa}\frac{\mu_2^\sigma(\kappa)^2}{1+V(\kappa)}\\
\hat{q}^2_\sigma=\alpha \mathbb{E}_{\kappa}\mu_2^\sigma(\kappa)^2\frac{(f^\star(\kappa)-m(\kappa))^2+q(\kappa)}{\left(1+V(\kappa)\right)^2}\\
\hat{\zeta}_\sigma=\alpha \mathbb{E}_{\kappa}\mu_1^\sigma(\kappa)\kappa\frac{(f^\star(\kappa)-m(\kappa))^2+q(\kappa)}{\left(1+V(\kappa)\right)^2}
\end{cases}
\end{align}
Importantly, these equations are conjectured to be asymptotically exact, but note that they are not fully asymptotic, in the sense that they still involve traces of large-dimensional matrices. Finally, the test error admits the compact asymptotic expression
\begin{align}
    \epsilon_g&=\mathbb{E}_{\kappa}\left[\left(
    \sigma_\star(\kappa)-\sum\limits_{\sigma\in V}\left[\mu_0^\sigma(\kappa)m_\sigma+\mu_1^\sigma(\kappa)\kappa\zeta_\sigma\right]
    \right)^2 +q(\kappa)\right]\notag\\
    &=\mathbb{E}_{\kappa}\left[\left(
    \sigma_\star(\kappa)-m(\kappa)
    \right)^2 +q(\kappa)\right]
\end{align}

\subsection{The special case $\W=\Pi^\perp$} 
\begin{figure}
    \centering
    \includegraphics[scale=0.65]{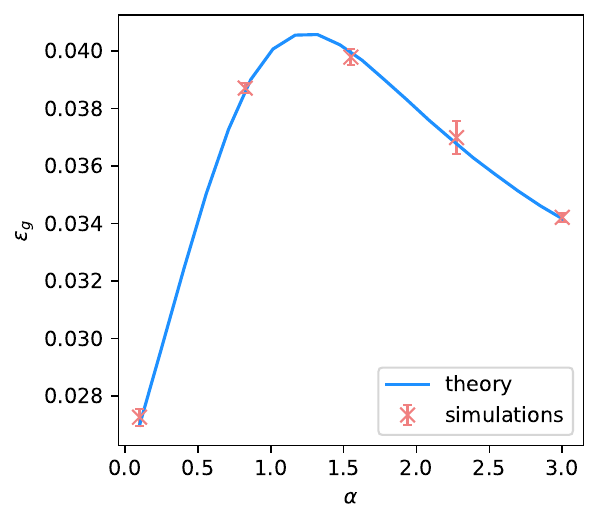}
    \caption{Test error for a sRF with $c=1, r=1, \gamma=1$, and ReLU activation for non-uniform $u$ in the special case $\W=\Pi^\perp$, when learning from a single-index model with $\sigma_\star=$ReLU. A regularization $\lambda=0.1$ is used to train the readout. The components of $u$ were drawn as independent Rademacher variables, $u_i\sim\mathrm{Rad}(\sfrac{1}{2})=\sfrac{1}{2}\delta_{\cdot,1}+\sfrac{1}{2}\delta_{\cdot,-1}$. Solid line: theoretical prediction of \eqref{eq: simplified_nonunif}, for $W=\Pi^\perp$. Crosses represent numerical experiments in dimension $d=2000$, averaged over $10$ runs. }
    \label{fig:non_uniform}
\end{figure}

We briefly discuss a special case of technical interest, for which the saddle-point equations admit a fully asymptotic formulation, namely when $\W=\Pi^\perp$. Then
\begin{align}
    \Omega_{\sigma\tau}\approx\delta_{\sigma\tau}\Pi_\sigma
\end{align}
and the saddle point equations reduce to
\begin{align}
\label{eq: simplified_nonunif}
    &\begin{cases}
        q_{\sigma}=n_\sigma \frac{\hat{q}_{\sigma}
}{\left(\hat{V}_{\sigma}+\lambda 
\right)^{2}}
    \\
V_{\sigma}=n_\sigma \frac{1}{\hat{V}_{\sigma}+\lambda 
}\\
m_\sigma=\frac{1}{\mathbb{E}_\kappa\left[\frac{\mu_0^\sigma(\kappa)^2}{1+V(\kappa)}\right]}
\mathbb{E}_\kappa\left[\frac{\mu_0^\sigma(\kappa)(f^\star(\kappa)-\sum\limits_{\tau\ne \sigma}\mu_0^\tau(\kappa)m_\tau)}{1+V(\kappa)}\right]
    \end{cases}
    \\
&\begin{cases}
\hat{V}_{\sigma}=\alpha \mathbb{E}_{\kappa}\frac{\mu_1^\sigma(\kappa)^2+\mu_2^\sigma(\kappa)^2}{1+V(\kappa)}\\
\hat{q}_\sigma=\alpha \mathbb{E}_{\kappa}(\mu_1^\sigma(\kappa)^2+\mu_2^\sigma(\kappa)^2)\frac{(f^\star(\kappa)-m(\kappa))^2+q(\kappa)}{\left(1+V(\kappa)\right)^2}
\end{cases}
\end{align}
Where $n_\sigma$ is the fraction of components of $u$ taking value $\sigma$. Fig.\,\ref{fig:non_uniform} represents the theoretical prediction of \eqref{eq: simplified_nonunif}, and displays good agreement with the corresponding numerical experiments.

Note that the formulae \eqref{eq: simplified_nonunif} are further amenable to being rewritten in functional form, denoting $\nu(x)$ the probability density of the components of $u$. One can then conjecture in this limit the following limiting equations
\begin{align}
    &\begin{cases}
        q(x)=\nu(x) \frac{\hat{q}(x)
}{\left(\hat{V}(x)+\lambda 
\right)^{2}}
    \\
V(x)=\nu(x) \frac{1}{\hat{V}(x)+\lambda 
}\\
m(x)=\frac{1}{\mathbb{E}_\kappa\left[\frac{\mu_0(x,\kappa)^2}{1+V(\kappa)}\right]}
\mathbb{E}_\kappa\left[\frac{\mu_0(x,\kappa)(f^\star(\kappa)-\int dz \nu(z)\mu_0(z,\kappa)m(z)}{1+V(\kappa)}\right]
    \end{cases}
    \\
&\begin{cases}
\hat{V}(x)=\alpha \mathbb{E}_{\kappa}\frac{(\mu_2(x,\kappa)^2+\mu_1(x,\kappa)^2)}{1+V(\kappa)}\\
\hat{q}(x)=\alpha \mathbb{E}_{\kappa}(\mu_2(x,\kappa)^2+\mu_1(x,\kappa)^2)\frac{(f^\star(\kappa)-m(\kappa))^2}{\left(1+V(\kappa)\right)^2}
\end{cases}
\end{align}
With the notations
\begin{align}
    &q(\kappa)=\int dx \nu(x) (\mu_1(x,\kappa)^2+\mu_2(x,\kappa)^2) q(x)\\
    &V(\kappa)=\int dx \nu(x) (\mu_1(x,\kappa)^2+\mu_2(x,\kappa)^2) V(x).
\end{align}
As discussed in the main text, the exploration of the case of non-uniform $u$ is of interest, as it affords a richer functional basis to learn the target $\sigma_\star$ (see also discussion below \eqref{eq:lower_bound}). This study falls well beyond the scope of the present work, and is left for future work.

\paragraph{A brief comment on incorporating bias} As mentioned in Appendix \ref{sec:app:replica}, we can accommodate the case of non-uniform quenched bias in the analysis, following identical lines as the one presented above, when the components of the bias can take values in a finite vocabulary $V_b$. Note that the initial step of the computation involves partitioning the indices $1\le i\le d$ as a function of the values of $u_i$, regrouping $i$s corresponding to the same values of $u_i=\sigma$ into the vectors $1_\sigma$ and matrices $\Omega_\sigma$. If non-uniform biases are present, identical conditioning should be done, this time regrouping terms corresponding to the same \textit{joint} value $\sigma,b$ of $u_i, b_i$. One implications is that instead of having $\mu^{\sigma}_{0,1,2}(\kappa)$ as above, one needs to consider $\mu^{\sigma,b}_{0,1,2}(\kappa)$. The analysis proceeds identically. At the level of the network expressivity, quenched but non-uniform biases allow the network to express functions from an even richer basis $\{\mu_0(\omega \cdot+b), \tilde{\mu}_1(\omega +b\cdot)\}_{(b,\omega)\in \mathcal{V}}$, where $\mathcal{V}\subset V_b\times V $ correspond to all realized pairs $b,\omega$ by $(b_i,u_i)$.

%% file: sections/appendix/Bounds.tex
\section{Derivation of the Bounds \eqref{eq:upper_bound} and \eqref{eq:lower_bound}}

\label{app:bounds}

\begin{figure}
    \centering
    \includegraphics[scale=0.4]{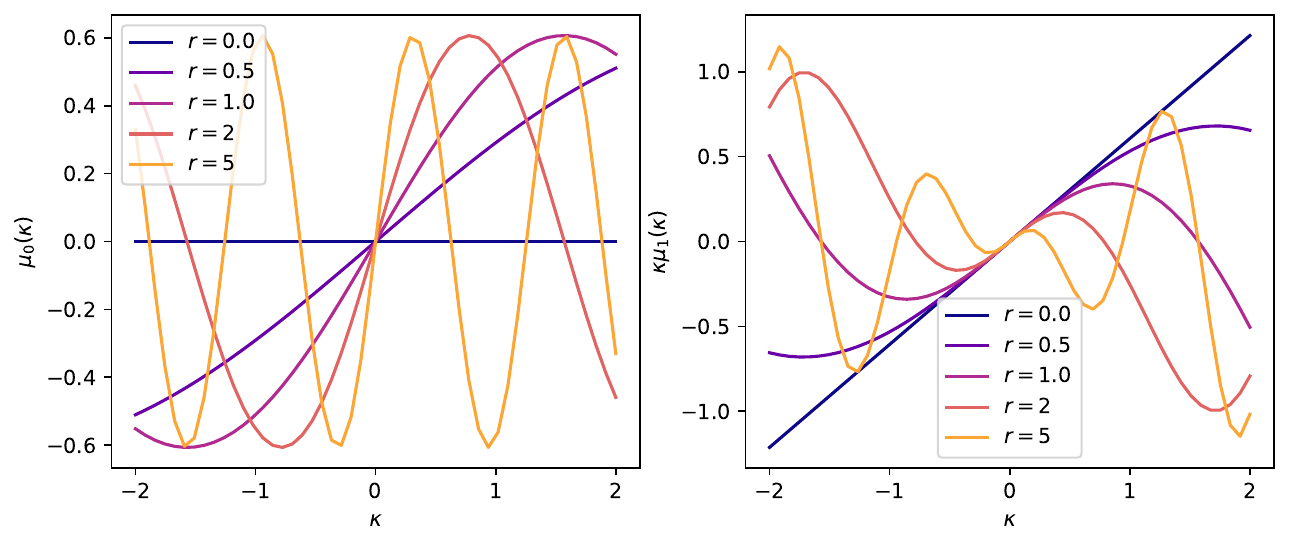}
    \includegraphics[scale=0.4]{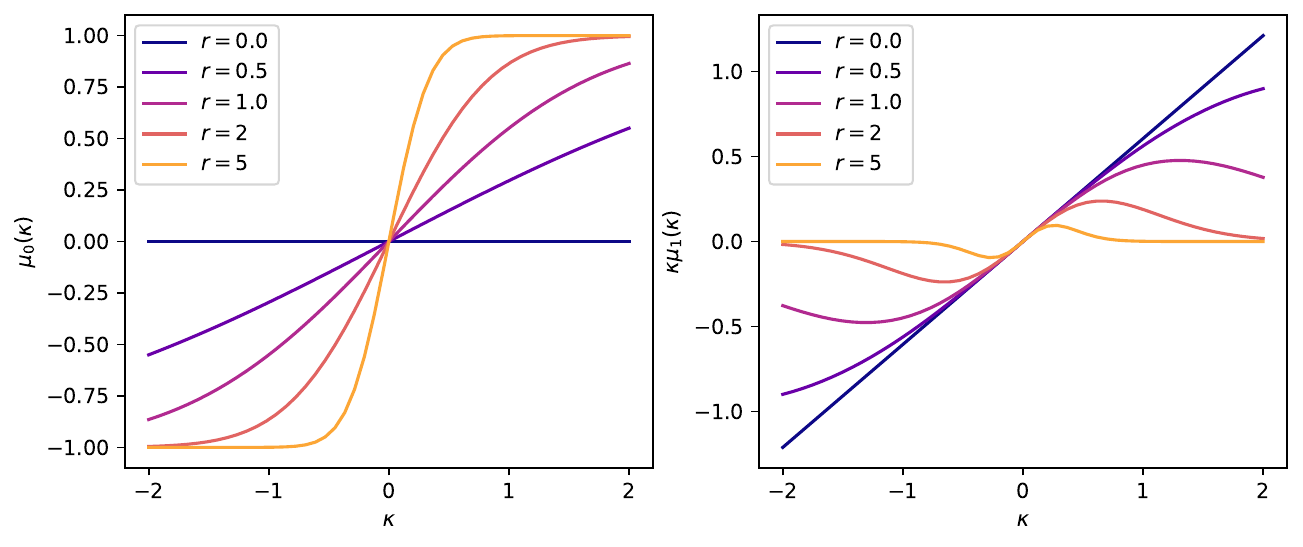}
    \caption{Vizualization of the functions $\mu_0(r), \Tilde{\mu}_1(r)$ for $\sigma=\tanh$ (bottom) and $\sigma=\sin$ (top), for various parameters $r$, in the perfectly aligned case $\gamma=1, c=1$, as can be achieved for $\alpha_0\to\infty$, see also discussion below equation \eqref{eq:lower_bound} in the main text. We remind that the spike strength $r$ cna be tuned via the learning rate $\Tilde{\eta}$. The functions $\mu_0(r), \Tilde{\mu}_1(r)$ provide the basis of functions that can be used to approximate the target $\sigma_\star$ . }
    \label{fig:mus_illustration}
\end{figure}

We detail -- and illustrate-- in this Appendix the derivation of the bounds \eqref{eq:upper_bound} and \eqref{eq:lower_bound} on the test error. Again, for definiteness, we consider the case of perfect spike/target alignment $\gamma=1$ (as can be achieved e.g. by using a large amount of data $\alpha_0\to\infty$ in the first step). 

\subsection{Derivation of the upper bound \eqref{eq:upper_bound}}

We first focus on the upper bound \eqref{eq:upper_bound}, showing that it can be attained for $\lambda \to\infty$. In this limit, equations \eqref{eq:SP} imply that all order parameters but $m$ are vanishing $q_1=q_2=V_1=V_2=\zeta=\psi=0$. Then the expression for $m$ simplifies as 
\begin{align}
\label{eq:m_as_proj}m=\frac{\mathbb{E}_\kappa[\mu_0(\kappa)\sigma_\star(\kappa)]}{\mathbb{E}_\kappa[\mu_o(\kappa)^2]}=\frac{\langle \mu_0, \sigma_\star\rangle}{\langle \mu_0, \mu_0\rangle}, 
\end{align}
where for two univariate functions $f,g$ we denoted the scalar product with respect to the Gaussian measure:
\begin{align}
\label{eq:scalar_product}
    \langle f,g\rangle\equiv \int \frac{1}{\sqrt{2\pi}}e^{-\sfrac{x^2}{2}}f(x)g(x).
\end{align}
In other terms, $m$ is proportional to the projection of the target function $\sigma_\star$ on $\mathrm{span}(\mu_0)$. Note that $m$ is can also be rewritten as
\begin{align}
    m=\underset{b_1}{\mathrm{arg inf}}\lVert 
    \sigma_\star-
    b_1\mu_0
    \lVert^2,
\end{align}
where the norm is with respect to the scalar product  \eqref{eq:scalar_product}. Finally, from \eqref{eq:test_error_replica}, the test error in the $\lambda\to\infty$ limit admits the compact expression
\begin{align}
    \underset{\lambda\to\infty}{\lim}\epsilon_g=\mathbb{E}_\kappa [(\sigma_\star(\kappa)-m\mu_0(\kappa))^2]=\underset{b_1}{\inf }\lVert 
    \sigma_\star-
    b_1\mu_0
    \lVert^2.
\end{align}
Finally, it follows that
\begin{align}
    \underset{\lambda \le 0}{\epsilon_g}\le\underset{\lambda\to\infty}{\lim}\epsilon_g=\underset{b_1}{\inf }\lVert 
    \sigma_\star-
    b_1\mu_0
    \lVert^2,
\end{align}
which recovers the upper bound \eqref{eq:upper_bound}.

\subsection{Derivation of the lower bound \eqref{eq:lower_bound}}

We now turn to the lower bound \eqref{eq:lower_bound}. First observe that by the definitions of $q_1, \psi,\rho$ \eqref{eq:summary_statistics}, it follows from the Cauchy-Schwarz inequality that
\begin{align}
    \psi\le \sqrt{q_1\rho}.
\end{align}
Therefore, for all $\kappa$
\begin{align}
    q(\kappa)-\frac{\mu_1(\kappa)^2\psi^2}{\rho}&=\mu_1(\kappa)^2\frac{\rho q_1-\psi^2}{\rho}+\mu_2(\kappa)^2 q_2 \ge 0
\end{align}
as a consequence, the test error \eqref{eq:test_error_replica} is lower bounded as
\begin{align}
    \epsilon_g&=\mathbb{E}_{\kappa}\Bigg[\Bigg(
    \sigma_\star(\kappa )-\mu_0(\kappa)m-\mu_1(\kappa)\kappa\zeta
    \Bigg)^2 {+q(\kappa)-\frac{\mu_1(\kappa)^2\psi^2}{\rho} }\Bigg]\notag\\
    &\ge \mathbb{E}_{\kappa}\Bigg[\Bigg(
    \sigma_\star(\kappa )-\mu_0(\kappa)m-\mu_1(\kappa)\kappa\zeta
    \Bigg)^2\Bigg]\notag\\
    &\ge \underset{b_1,b_2}{\inf}\mathbb{E}_{\kappa}\Bigg[\Bigg(
    \sigma_\star(\kappa )-\mu_0(\kappa)b_1-\mu_1(\kappa)\kappa b_2
    \Bigg)^2\Bigg], 
\end{align}
which recovers the lower bound \eqref{eq:lower_bound}. The functions corresponding to the upper (resp. lower) bound (see \eqref{eq:upper_bound}, \eqref{eq:lower_bound}) are represented for a $\sigma_\star=\sin$ target in Fig.\,\ref{fig:bounds}.

\subsection{Derivation of bounds \eqref{eq:bound_opteta}}

Since the bounds \eqref{eq:upper_bound} and \eqref{eq:lower_bound} hold for all learning rates $\Tilde{\eta}$, the bounds \eqref{eq:bound_opteta} follow immediately from taking the minimum over $\Tilde{\eta}$ on the right-hand side and left-hand side of \eqref{eq:upper_bound} and \eqref{eq:lower_bound}. The functions $\{\mu_0(r),\Tilde{\mu}_1(r)\}_{r\ge 0}$, which form the pool of functions which can be used to approximate the target function, are represented for $\sigma=\sin,\tanh$ in Fig.\,\ref{fig:mus_illustration}.

%% file: sections/appendix/Spectrum.tex
\section{Spectrum of The Features Empirical Covariance Matrix}

\begin{figure}
    \centering
    \includegraphics[scale=0.5]{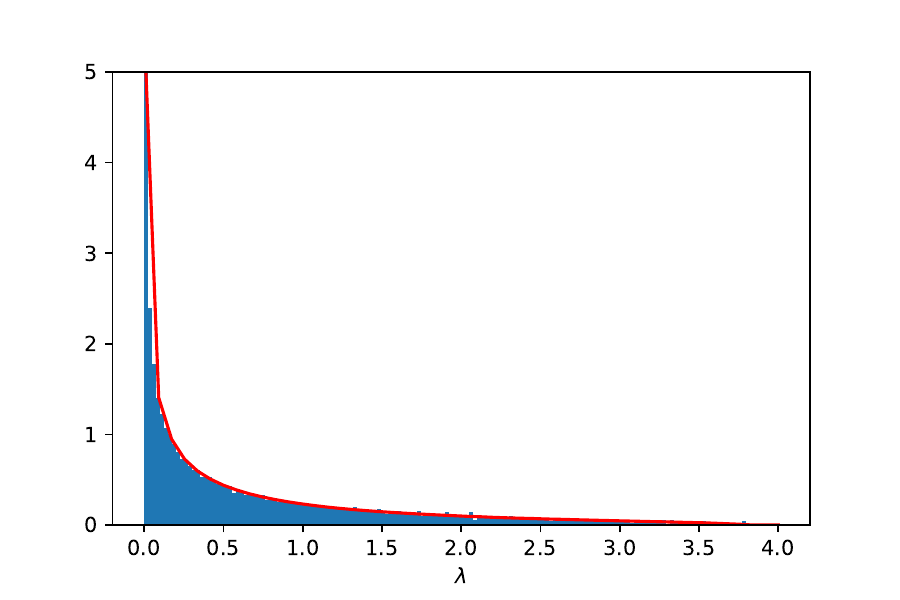}
    \caption{Spectrum of the empirical features covariance for $\alpha=1,\beta=1,\sigma=\mathrm{sign}$. Blue: numerical estimation in $d=2000$. Red: theoretical characterization \eqref{eq:app:stieltjes}. There is a single spike $\lambda_{\mathrm{spk.}}=\Theta_d(d)$ in addition to the bulk, which is not represented in this figure. }
    \label{fig:app:spectrum}
\end{figure}

\label{sec:app:Stieltjes}

In this section, we derive a tight asymptotic expression for the spectrum of the empirical covariance of the sRF features $\{\varphi(\vec x^\mu)\}_{\mu=1}^n$. Before starting the derivation, let us first remind that the \textit{population} covariance is given by the conditional Gaussian equivalence \ref{conj:main:gaussian_equivalence} as 
\begin{align}
\mathbb{E}_{\vec x}[\varphi(\vec x)\varphi(\vec x )^\top]=\mathbb{E}_\kappa\left[
\left(\mu_0(\kappa){\vec 1}_p+\frac{1}{\sqrt{d}}\kappa \mu_1(\kappa)W{\vec v}\right)^{\otimes 2}+\mu_1(\kappa)^2 W\Pi^\perp W^\top +\mu_2(\kappa)^2\mathbb{I}_p
\right]
\end{align}
where $\cdot^{\otimes 2}$ denotes the outer product of a vector with itself. \\

We now turn to the characterization of the empirical covariance. Let us denote by $\Psi\in \mathbb{R}^{n\times p}$ the matrix with rows  $\{\varphi(\vec x^\mu)\}_{\mu=1}^n$, and the empirical covariance $\Sigma=\sfrac{1}{p} \Psi^\top \Psi$. Note that up to zero eigenvalues, this matrix has the same spectrum as the conjugate kernel $\sfrac{1}{p}\Psi\Psi^\top$. In order to access the spectrum of $\Sigma$, we aim to compute the Stieltjes transform
\begin{align}
s(z)=\mathbb{E}_{\{\vec x^\mu\}_{\mu=1}^n}\Tr[(\Sigma+z\mathbb{I}_p)^{-1}]
\end{align}
for some complex number $z$. This computation can be achieved again using the replica method, first observing the identity
\begin{align}
s(z)= -2\partial_z \mathbb{E}_{\{\vec x^\mu\}_{\mu=1}^n}\frac{1}{2}\ln \det[2\pi(\Sigma+z\mathbb{I}_p)^{-1}]\equiv 2 \partial_z\Phi(z).
\end{align}
We introduced the free entropy $\Phi(z)$, which can be computed as
\begin{align}
\Phi(z)&=\mathbb{E}_{\{\vec x^\mu\}_{\mu=1}^n}\ln \underbrace{ \int d\vec{w} e^{-\frac{1}{2}\vec{w}^\top (\Sigma+z\mathbb{I}_p) \vec{w}}}_{\equiv Z(z)}.
\end{align} 
Once more, the generative function $\mathbb{E}_{\{\vec x^\mu\}_{\mu=1}^n} \ln Z(z)$ can be evaluated using the replica method, leveraging the identity
\begin{align}
 \mathbb{E}_{\{\vec x^\mu\}_{\mu=1}^n}\ln Z(z)=\underset{s\to 0}{\lim}\frac{\ln (\mathbb{E}_{v} Z(z)^s)-1}{s}.
\end{align}
The problem thus reduces to the computation of the replicated partition function $\mathbb{E}_{\vec x} Z(z)^s$
\begin{align}
\mathbb{E}_{\{\vec x^\mu\}_{\mu=1}^n}  Z(z)^s&= \int \prod\limits_{a=1}^s d \vec{w}_a e^{-\frac{z}{2}\sum\limits_a \lVert \vec{w}_a\lVert^2} \prod\limits_{\mu}\left[\mathbb{E}_{\vec{x}}
e^{-\frac{1}{2}\sum\limits_a(\sfrac{\vec{w}^\top_a \varphi(\vec{x})}{\sqrt{p}})^2  }
\right]
\notag\\
&= \int \prod\limits_{a=1}^s d \vec{w}_a e^{-\frac{z}{2}\sum\limits_a \lVert \vec{w}_a\lVert^2} \left[\mathbb{E}_\kappa \mathbb{E}_{\lambda }
e^{-\frac{1}{2}\sum\limits_a (\mu_0(\kappa)m_a+\mu_1(\kappa)\kappa \zeta_a +\lambda_a)^2  }.
\right]^n
\end{align}
Like in Appendix \ref{sec:app:replica}, we introduced the order parameters
\begin{align}
m_a\equiv\frac{\vec{w}_a^\top \vec{1}_p}{\sqrt{p}}, && \zeta_a\equiv\frac{\vec{w}_a^\top W \vec{v}}{\sqrt{dp}}.
\end{align}
We also introduced the local fields $\{\lambda_a\}_a$, which are Gaussian with zero mean and joint statistics
\begin{align}
\mathbb{E}[\lambda_a\lambda_b]=\mu_1(\kappa)^2 q^1_{ab}+\mu_2(\kappa)q^2_{ab},
\end{align}
denoting the summary statistics
\begin{align}
q^1_{ab}\equiv\frac{\vec{w}_a^\top \Omega \vec{w}}{p}, &&q^2_{ab}\equiv \frac{\vec{w}_a^\top \vec{w}}{p},
\end{align}
where we remind $\Omega=W\Pi^\perp W^\top$. The replicated partition function can thus be written as
\begin{align}
    Z^s=&\int \prod\limits_{a=1}^s dm_ad\hat{m}_ad\zeta_ad\hat{\zeta}_a\prod\limits_{a,b}dq^1_{ab}d\hat{q}^1_{ab}
    dq^2_{ab}d\hat{q}^2_{ab}\underbrace{ e^{-\sqrt{p}\sum\limits_r m_a\hat{m}_a
    -\sqrt{dp}\sum\limits_a(\zeta_a\hat{\zeta}_a)
    -p\sum\limits_{a\le b}(q^1_{ab}\hat{q}^1_{ab}+q^2_{ab}\hat{q}^2_{ab})}}_{e^{s p \Psi_t}}
    \notag\\
    &\underbrace{\int \prod\limits_{a}d \vec w_a e^{-\frac{z}{2} \sum\limits_a \lVert\vec w_a\lVert^2}e^{+\sum\limits_{a=1}^s \vec w_a^\top(\hat{m}_a  \vec 1_p+\hat{\zeta}_a \W \vec v ) +\sum\limits_{a\le b} (\hat{q}^1_{ab} \vec w_a^\top \Omega \vec w_{b}+\hat{q}^2_{ab} \vec w_a^\top \vec w_b)}}_{e^{s  p \Psi_w}}\notag\\
    &\underbrace{\left[\mathbb{E}_{\kappa}\left(\mathbb{E}_{\{\lambda_a\}_{a=1}^s} e^{-\frac{1}{2}\sum\limits_{a=1}^s (\mu_0(\kappa) m_a+\mu_1(\kappa)\kappa \zeta_a+\lambda_a)^2}\right)\right]^n}_{e^{\sfrac{\alpha}{\beta} s p\Psi_y}},
\end{align}
an expression bearing great similarity with that reached in Appendix \ref{sec:app:replica}. In fact, formally, this expression is equivalent to the finite-temperature ($\omega=1$) version of the one discussed in Appendix \ref{sec:app:replica}, in the particular case of a regularization $\lambda=z$, and with a loss function $\ell(y, z)=\ell(z)=z^2$. Identical computational steps can then be followed to reach a tight characterization of the free entropy. Again, we assume a RS ansatz
\begin{align}
&\forall 1\le a,b\le s, ~~q^1_{ab}=\delta_{ab}V_1\notag\\
&\forall 1\le a,b\le s, ~~q^2_{ab}=\delta_{ab}V_2\notag\\
&\forall 1\le a\le s, ~~\zeta_a=\zeta\notag\\
&\forall 1\le a\le s, ~~m_a=m\notag\\
&\forall 1\le a,b\le s, ~~\hat{q}^1_{ab}=-\delta_{ab}\sfrac{\hat{V}_1}{2}\\
&\forall 1\le a,b\le s, ~~\hat{q}^2_{ab}=-\delta_{ab}\sfrac{\hat{V}_2}{2}\\
&\forall 1\le a\le s, ~~\hat{\zeta}_a=\hat{\zeta}\notag\\
&\forall 1\le a\le s, ~~\hat{m}_a=\hat{m}.
\end{align}
Following identical steps as Appendix \ref{sec:app:replica}, one finally reaches
\begin{align}
\Phi(z)=&\underset{V_1,V_2,m,\zeta,\hat{V}_1,\hat{V}_2,\hat{m},\hat{\zeta}}{\mathrm{extr}}
\frac{\hat{V}_1V_1+\hat{V}_2V_2}{2}-\frac{m\hat{m}}{\sqrt{p}}-\frac{\zeta\hat{\zeta}}{\sqrt{\beta}}-\frac{1}{2}\ln\det\left[z\mathbb{I}_p+\hat{V}_1\Omega+\hat{V}_2\mathbb{I}_p\right]\notag\\
&
+\frac{1}{2}\Tr[(z\mathbb{I}_p+\hat{V}_1\Omega+\hat{V}_2\mathbb{I}_p)^{-1}\left(
\hat{m}^2\vec{1}_p\vec{1}_p^\top +\hat{\zeta}^2 W\vec{v}\vec{v}^\top W^\top
\right)]
\notag\\
&
-\frac{\alpha}{\beta}\frac{1}{2}\mathbb{E}_{\kappa}\left[\frac{\left(\mu_0(\kappa)m+\mu_1(\kappa)\kappa \zeta\right)^2}{1+V(\kappa)}\right]-\frac{\alpha}{\beta}\frac{1}{2}\mathbb{E}_{\kappa}\ln(1+V(\kappa))
\end{align}
Expunding the extremization conditions with respect to $m,\hat{m}, \zeta, \hat{\zeta}$ imposes
\begin{align}
\begin{cases}
0=\hat{m}\Tr[(z\mathbb{I}_p+\hat{V}_1\Omega+\hat{V}_2\mathbb{I}_p)^{-1}
\vec{1}_p\vec{1}_p^\top ]\\
\zeta=\hat{\zeta}\Tr[(z\mathbb{I}_p+\hat{V}_1\Omega+\hat{V}_2\mathbb{I}_p)^{-1}
W\vec{v}\vec{v}^\top W^\top ]\\
m=-\frac{\mathbb{E}_\kappa\left[\frac{\mu_1(\kappa)\kappa \mu_0(\kappa)}{1+V(\kappa)}\right]}{\mathbb{E}_\kappa\left[\frac{\mu_0(\kappa)^2}{1+V(\kappa)}\right]}\zeta\\
\hat{\zeta}=\mathbb{E}_\kappa\left[\frac{\left(\mu_0(\kappa)m+\mu_1(\kappa)\kappa \zeta\right)\mu_1(\kappa)\kappa}{1+V(\kappa)}\right]
\end{cases}
\end{align}
Combining the last two equations imposes
\begin{align}
\hat{\zeta}=\mathbb{E}_\kappa\left[\frac{\left(-\mu_0(\kappa)\frac{\mathbb{E}_\kappa\left[\frac{\mu_1(\kappa)\kappa \mu_0(\kappa)}{1+V(\kappa)}\right]}{\mathbb{E}_\kappa\left[\frac{\mu_0(\kappa)^2}{1+V(\kappa)}\right]}+\mu_1(\kappa)\kappa \right)\mu_1(\kappa)\kappa}{1+V(\kappa)}\right]\zeta
\end{align}
from which it generically follows that $\zeta=0$, which also implies $m=\hat{m}=\hat{\zeta}=0$. Finally, the extremization conditions with respect to $V_{1,2}, \hat{V}_{1,2}$ yield the simple equations
\begin{align}
\begin{cases}
V_2=\Tr[(z\mathbb{I}_p+\hat{V}_1\Omega+\hat{V}_2\mathbb{I}_p)^{-1}]\\
V_1=Tr[(z\mathbb{I}_p+\hat{V}_1\Omega+\hat{V}_2\mathbb{I}_p)^{-1}\Omega]\\
\hat{V}_1=\frac{\alpha}{\beta}\mathbb{E}_\kappa\left[\frac{\mu_1(\kappa)^2}{1+V(\kappa)}\right]\\
\hat{V}_2=\frac{\alpha}{\beta}\mathbb{E}_\kappa\left[\frac{\mu_2(\kappa)^2}{1+V(\kappa)}\right]
\end{cases}
\end{align}

Note how these equations correspond exactly to the ones of Result \ref{res:asymptotics}, replacing the regularization $\lambda$ by the Stieltjes argument $z$. Finally, we observe that $\partial_z\Phi(z)=\sfrac{1}{2}V_2$, which implies that the Stieltjes transform $s(z)$ is compactly given by 
\begin{align}
\label{eq:app:stieltjes}
s(z)=V_2,
\end{align}
with $V_2$ a solution of the system of equations above. The spectrum can then be extracted from the Stieltjes transform, and matches well with numerical simulations on sRF features, see Fig.\,\ref{fig:app:spectrum}.

\paragraph{Spike} The previous study allowed to characterize the asymptotic bulk eigenvalue distribution of the empirical features covariance. Note that there is a single spike $\lambda_{\mathrm{spk.}}=\Theta_d(d)$ not captured by the analysis, which relies on a Stieltjes argument $z=\Theta_d(1)$. It is however reasonable to expect this spike eigenvalue to coincide with that of the population covariance. This eigenvalue should then be asymptotically equal to
\begin{align}
    \frac{\lambda_{\mathrm{spk.}}}{d}=\alpha \mathbb{E}_{\kappa\sim\mathcal{N}(0,1)}[c_0(\kappa)^2].
\end{align}